\documentclass[natbib,doc,nolmodern]{apa6}

\usepackage[english]{babel}
\usepackage{amsmath,amssymb}
\usepackage{txfonts}  %%% fixes problem with \boldsymbol
\usepackage{graphicx}
\usepackage{pgf}
\usepackage{pgfplots}
\usepackage{tikz}
\usepackage{xcolor}
\usepackage[colorlinks=true,allcolors=blue]{hyperref}
\usepackage{linguex}
\usepackage{multirow}
\usepackage{dcolumn}
\usepackage{rotating}
\usepackage[colorinlistoftodos,prependcaption,textsize=scriptsize]{todonotes}

\usepackage{colortbl}
\usepackage{hyperref}

\newcommand{\changedSecond}[1]{#1}
\newcommand{\changed}[1]{#1}
\newcolumntype{d}{D{.}{.}{3.2}}
\newcolumntype{i}{D{.}{.}{1.3}}

\newcommand{\mc}{\multicolumn}

\def\addlegendimage{\csname pgfplots@addlegendimage\endcsname}

\pgfkeys{/pgfplots/number in legend/.style={%
        /pgfplots/legend image code/.code={%
            \node at (0.125,-0.0225){#1}; % <= changed x value
        },%
    },
}
\pgfplotsset{
every legend to name picture/.style={west}
}

\newcommand{\citegen}[1]{\citeauthor{#1}'s (\citeyear{#1})}

\title{Modeling Task Effects in Human Reading with\\[0.8ex]
  Neural Network-based Attention}

\shorttitle{Modeling Task Effects in Human Reading}

\twoauthors{Michael Hahn}{Frank Keller}

\twoaffiliations{%
Department of Linguistics\\
Stanford University\\
\url{mhahn29@gmail.com}}{%
School of Informatics\\
University of Edinburgh\\
\url{keller@inf.ed.ac.uk}}

\leftheader{Hahn and Keller}

\abstract{Research on human reading has long documented that reading behavior shows task-specific effects, but it has been challenging to build general models predicting what reading behavior humans will show in a given task. We introduce NEAT, a computational model of the allocation of attention in human reading, based on the hypothesis that human reading optimizes a tradeoff between economy of attention and
success at a task. Our model is implemented using contemporary neural network
modeling techniques, and makes explicit and testable predictions about how the
allocation of attention varies across different tasks. We test this in an eyetracking
study comparing two versions of a reading comprehension task, finding that our
model successfully accounts for reading behavior across the tasks. Our work thus
provides evidence that task effects can be modeled as optimal adaptation to task
demands.}

\authornote{Preprint of a paper accepted for publication at \emph{Cognition}.
A preliminary version of the modeling study reported in
  Section~\ref{sec:study1} was previously published in
  \citet{Hahn:Keller:16}. For the present article, the model has been
  reimplemented and all modeling experiments have been rerun and the
  results reanalyzed. Data and code are freely available at
  \url{https://gitlab.com/m-hahn/task-effects-neural-networks}.}

\begin{document}

\maketitle

%\newpage
%\tableofcontents

%%%%%%%%%%%%%%%%%%%%%%%%%%%%%%%%%%%%%%%%%%%%%%%%%%%%%%%%%%%%%%%%%%%%%%%%%%%%%%%%
\section{Introduction}
\label{sec:intro}

Explaining the cognitive processes involved in human reading poses a
particularly interesting challenge for cognitive science. First,
because reading is a cultural skill that is acquired through explicit
instruction combined with a large amount of practice. This sets it
apart from other linguistic skills such as speaking and listening,
which are normally acquired earlier in life than reading and do not
require explicit instruction. Often, an innate component is assumed to
be part of language acquisition; but this is not a plausible
assumption for reading, as it is a recent phenomenon by evolutionary
standards, dating back to around 3000--4000~BC. Reading, insofar as it
is distinct from the rest of linguistic cognition, therefore must be a
skill that is learnable from experience.

A second important aspect of human reading is that it is a highly
task-specific process. When we read a book for entertainment, we will
do so differently than when we read the same book in order to find
typographical errors. For many tasks, readers employ specific
information-seeking strategies. For example, to find an answer to a
question in a text, they systematically search for names, dates,
amounts, or whatever form the correct answer is likely to
take. Reading strategy will also vary depending on whether the task is
to memorize facts (perhaps for an exam), translate a text, compile a
summary of book, or write a review. It is an intriguing question how
the high-level cognitive requirements of such tasks are translated
into low-level reading behavior, which ultimately manifests itself as
a sequence of eye-movements on the words of a text.

Both the experiential nature of reading and its task specificity pose
challenges for computational cognitive modeling: on the one hand, we
need to develop a model that assumes only a minimum of innate
knowledge and is able to learn the key properties of human reading
from exposure to large amounts of text. On the other hand, our model
must be amenable to explicit instruction (perhaps in the form of
feedback) and has to be capable of learning different reading
strategies for different tasks. In other words, we expect the model to
change its eye-movement behavior depending on whether its task is to
memorize information, answer a question, find typos in a text,~etc.

In this article, we focus on one particular aspect of reading, viz.,
the allocation of attention. We assume that the allocation of
attention can be studied by measuring the eye-movements that humans
make as they read. These eye-movements consist of fixations and
saccades: during a fixation, the eyes land on a word and remain fairly
static for 200--250~ms. Saccades are the rapid jumps that occur
between fixations, typically lasting 20--40~ms and spanning
7--9~characters \citep{rayner_eye_1998}. Readers, however, do not
simply fixate one word after another; some saccades go in reverse
direction, some words are fixated more than once, and other words are
skipped altogether.

We present a computational cognitive model that captures the
allocation of attention during reading by modeling skipping, i.e., the
process that decides which words should be fixated, and which ones
should be skipped, by the reader. We will assume that a fixation
strategy can be learned from large amounts of text if an explicit task
is given to the reader. The task (such as text reconstruction or
question answering) allows the model to infer which words are
important for the task (and should be attended to), and which ones are
less important (and should be skipped). Our model also includes a
component which predicts how expected a word is given its context and
a series of prior fixations and skips. This component allows the model
to predict reading time for words in a text. Therefore, our model not
only computes which words are attended to in a text, but also how much
attention the fixated words receive.

When developing our model, we make the assumption that task-based
reading behavior can be explained by a fundamental tradeoff: the model
needs to trade off economy of attention (skipping as many words as
possible, i.e., reading as fast as possible) and accuracy (making as
few errors as possible in the task the reader wants to accomplish).
Task-specific reading strategies emerge when the model learns the
economy--accuracy tradeoff for a given task. Each task potentially
requires a different tradeoff, thus occasioning task-specific
differences in reading behavior.

As already mentioned, reading acquisition in humans typically involves
explicit instruction. For example, at school, children may have to
answer comprehension questions to determine if they have read a text
correctly. We assume that our model receives a similar type of
feedback in order to enable it to learn the economy--accuracy tradeoff
for a specific reading task.  This feedback comes in the form of a
reinforcement signal, i.e., information regarding whether the model
has accurately solved the task for a given input or not. While the
reinforcement signal is a type of instruction, it does not mean that
we assume supervised learning, i.e., that the model is told explicitly
which words to skip and to fixate, or for how long. This form of
supervision would not be cognitively plausible: humans who are taught
to read do receive feedback regarding whether they performed their
reading task correctly or not, but they are not told whether they have
fixated the right words, or spent the right amount of time on each
word.

In the following, after an overview of background and related work, we
will introduce the Tradeoff Hypothesis that underlies our model in
more detail. Then we will present the model itself, which is
implemented as an attention-based recurrent neural network that
performs word prediction under uncertainty (i.e., it can decide to
skip words). We derive its objective function for the baseline task of
reconstructing the input text, and show how the model can be trained
using reinforcement learning techniques. In Modeling Study~1, we
demonstrate that our model successfully captures skipping patterns and
reading times observed in a large eye-tracking corpus.

We then turn to a key prediction of our modeling approach: that
eye-movement behavior should change as a function of the task a reader
aims to solve. We present Experiment~1, an eye-tracking study designed
to test this prediction. This experiment uses question answering as
the task participants have to perform and manipulates whether they see
a preview of the question they have to answer before they read the
corresponding text. The results show a strong effect of task condition
(preview or not) on skipping rate, reading time, and task accuracy. We
also find evidence that participants adapt their reading strategy for
the words that are part of the potential or actual answer to the
question for a given text.

The results of Experiment~1 inform the design of an extended version
of our computational model, which now performs questions answering
instead of text reconstruction. In Modeling Study~2, we shows that
this extended model captures the reading behavior observed in
Experiment~1, predicting the effect of the preview manipulation on
skipping rate and task accuracy, while also correctly modeling the
reading strategy effects that we found experimentally.

The article concludes with a general discussion, which summarizes our
contributions, contextualizes them with respect to prior research, and
addresses some of the limitations of this work.

%%%%%%%%%%%%%%%%%%%%%%%%%%%%%%%%%%%%%%%%%%%%%%%%%%%%%%%%%%%%%%%%%%%%%%%%%%%%%%%% 
\section{Background}
\label{sec:related}

A significant literature exists in the domain of modeling human
reading. Here, we will review work on models of eye-movement control,
models of processing difficulty during language comprehension, and
more generally neural approaches to cognitive modeling, with a
particular focus on neural network-based attention mechanisms.

\subsection{Attention in Humans and Machines}
\label{sec:attention}

In this article, we will leverage recent developments in neural
network technology to develop a cognitively plausible model of human
reading that is able to process naturalistic text. More specifically,
we will use a mechanism in neural networks called attention, which
allows a model to dynamically focus on a restricted part of the
input. Attention is also a central concept in cognitive science, where
it denotes the focus of cognitive processing. In both language
processing and visual processing, attention is known to be limited to
a restricted area of the visual field; often it is assumed that
eye-movements indicate shifts in attention \citep{Henderson:03}. There
is therefore an obvious conceptual link between neural network-based
attention and human attention that we can exploit for model design.

\changed{In human vision, a distinction is sometimes made between
  overt and covert attention, where overt attention refers to the
  location that is behaviorally prominent a given point in time
  (typically the point of fixation), while covert attention refers to
  the current focus of cognitive processing. For example, it is
  possible that visual recognition occurs at a given location even
  though that location is not overtly attended (i.e.,~fixated), as
  \citet{Heyman:ea:17} show. Furthermore, there is evidence that
  viewers can classify visual scenes (i.e.,~perceive the gist of a
  scene) in the absence of overt attention
  \citep{Li2002RapidNS}. Findings of this sort indicate that covert
  and overt attention can dissociate in certain situations; however,
  other authors argue that overt and covert attention are so strongly
  related that studying them separately is not productive
  \citep[see][]{Henderson:03}.}

\changed{In the reading literature, a similar debate exists about
  whether the focus of cognitive processing and the overt focus of
  attention (the fixation point) can be dissociated. Here, the key
  distinction is between accounts of reading that assume strictly
  serial processing (one word after the other), and approaches that
  allow parallel processing (several words can be processed at the
  same time). Parallel processing implies a dissociation between
  covert attention (the processing of a word) and overt attention (the
  fixation of a word). We will discuss this in more detail in the next
  section.}

\subsection{Models of Eye-movement Control}
\label{sec:eye-movement_models}

Models of eye-movement control are designed to make detailed
predictions about the pattern of fixations and saccades that occur
when people read text. They typically integrate linguistic constraints
(such as lexical representations) with perceptual factors (such as
information about the length of a word or its visual appearance) to
compute the location and the duration of eye-movements during reading.
A range of computational models of eye-movement control have been
developed \citep[see][for an overview]{rayner_models_2010-1}, the most
prominent of which are probably E-Z Reader \citep{reichle_toward_1998,
  reichle_ez_2003, reichle_using_2009} and SWIFT
\citep{engbert_dynamical_2002, engbert_swift:_2005}.

E-Z Reader assumes that eye-movements are controlled in a serial
fashion: a single word is processed at any given time, and once
processing of that word if finished, attention moves on to the next
word. The model assumes two processing stages. The first one is the
familiarity check, which takes place during lexical access up to the
point when the word can be reliably identified. Once this point is
reached, the saccade to the next word is programmed, and the second
stage of processing, the completion of lexical access, occurs. Saccade
programming itself is also subdivided into two stages: the initial
labile stage, during which a saccade can still be canceled, and the
non-labile stage, when the saccade has become obligatory. Another key
assumption of E-Z Reader is that eye-movements target the center of
words, but are subject to random over- and undershoot errors; large
errors result in automatic refixations, i.e., corrective saccades
towards the center of the word.

SWIFT is a parallel model of eye-movement control. This means that
attention is conceptualized as a gradient that is spread across
several words, and lexical access for these words happens at the same
time.  Another key assumption of the model is that the decision to
move the eyes to the next viewing location is determined by a random
timer. This timer initiates saccades at random intervals; factors such
as word frequency influence saccade programming only indirectly by
inhibiting the timer. The inhibition delays when a saccade occurs,
resulting in an increased fixation duration on the current word.
SWIFT's assumptions about saccade programming are similar to the ones
made by E-Z Reader.

More recent approaches, such as the model by
\citet{bicknell_rational_2010-1} conceptualize reading as a Bayesian
inference task. The Bicknell and Levy approach assumes a probabilistic
language model which defines a prior distribution over possible word
sequences. The aim of the model is to decide whether or not to move
the eyes from the current position. While at a given position, it
samples noisy visual input, which results in a likelihood term (the
probability of the visual input given a word). By combining the prior
distribution with the likelihood term, the model updates its belief
about the identity of the word sequence it is reading. This updated
posterior distribution is then used to decide the next reading
action. \citet{bicknell_rational_2010-1} assume four possible actions:
keep fixating the current position, initiate a backward saccade,
initiate a forward saccade, or stop processing the current input. The
model uses a simple control policy based on the posterior probability
of the currently fixated character to decide which action to take. A
distinguishing factor of this model is that it assumes realistic
visual input, viz., noisy representations based on a character's
eccentricity from the viewing position.

Finally, there is work in the natural language processing literature
which treats eye-movement prediction as a machine learning task. More
specifically, these authors train models such as logistic regression
or conditional random fields on a corpus of human eye-tracking data,
and then predict fixation time, skipping rate, and other
eye-movement measures on an unseen test set.  Unlike our goals here,
the literature in this tradition
\citep[e.g.,][]{nilsson_learning_2009, nilsson_towards_2010,
  hara_predicting_2012, matthies_blinkers_2013, Hollenstein2021CMCL2S,
  Bestgen2021LASTAC} does not primarily aim to construct explanatory
cognitive models, and the use of supervised training (i.e.,~models
learn their behavior from a pre-existing training set of eye-movement
data) is not psychologically realistic, as outlined in the
introduction. However, machine learning-based models typically achieve
good prediction accuracy, which makes them suitable for comparison
with cognitive models.

The model proposed in this article does not aim to capture
eye-movement control at the same level of detail as E-Z Reader, SWIFT,
or the Bicknell and Levy model. We do not aim to predict where on a
word a fixation will land, or which distance a saccade will span. We
have no notion of reverse eye-movements or corrective saccades, and we
do not model the visual input \changedSecond{(except to a limited extent in the
pre-view component of our model)}. Instead, our model uses
an idealized, high-level notion of attention: it decides whether to
fixate or skip a word, and is also able to predict reading times for
fixated words (though this is merely a by-product of our model
architecture, as will be explained below).

On the other hand, our approach assumes an explicit tradeoff between
economy of attention and task accuracy. It shares this feature with
\citet{bicknell_rational_2010-1}, whose model trades off reading speed
(a form of economy) and visual identification accuracy. Also like
Bicknell and Levy, we assume that our model explicitly integrates
probability distributions from a variety of sources and models
decision making explicitly. We do not assume a Bayesian framework, but
compute probability distributions using a more flexible neural network
approach, and we capture decision making using reinforcement learning.
Again like Bicknell and Levy, our approach incorporates an explicit
language model, which in our case is computed using a recurrent neural
network.  However, whereas their model is designed specifically for
optimal word recognition, our Tradeoff Hypothesis generalizes to
higher-level language understanding tasks such as question answering.

\changed{Following in the footsteps of Bicknell and Levy,
  \citet{Lewis:ea:13} propose a model of saccade control that combines
  Bayesian optimization and bounded optimal control. Their model
  assumes an explicit speed-accuracy tradeoff which enables
  eye-movements to adapt to task conditions. They operationalize the
  tradeoff by assuming a set of predefined payoff schemes (focused on
  accuracy, focused on speed, or balanced). Their model is able to
  capture how human eye-movements in a list lexical decision task vary
  with the payoff scheme participants are given. The aim of our work
  is different from that of \citet{Lewis:ea:13} in that we develop
  explicit computational models across several tasks and derive the
  tradeoff between economy of attention and task accuracy from these
  models through reinforcement learning, rather than having to
  stipulate external payoff schemes.}

\subsection{Models of Processing Difficulty}
\label{sec:processing_diff_models}

Model of processing difficulty are designed to predict how much
processing effort is caused during sentence comprehension.
Experimental studies typically find that certain words in a sentence
cause increased processing effort, which empirically manifest itself
as greater reading time, but also in other ways, \changedSecond{for example slower
reaction times in a go/no go task or increased ERP amplitudes.}  Models
of processing difficulty are not designed to predict detailed
eye-movement behavior; they have no notion of fixations and saccades,
and do not model the visual properties of words. Often, they are
tested not on naturalistic texts, but on carefully designed sentence
pairs such as they are used in psycholinguistic experiments, garden
path sentences being a prime example.

A prominent model of processing difficulty is surprisal, which
measures the predictability of a word in context, defined as the
negative logarithm of the conditional probability of the current word
given the preceding words \citep{hale_probabilistic_2001,
  levy_expectation-based_2008}.  Surprisal is computed by a language
model, which can take the form of a probabilistic grammar, an n-gram
model, or a neural network. Like other models of processing
difficulty, surprisal is designed to provide a general measure of
linguistic processing effort, and cannot explain detailed eye-movement
behavior during reading. Nevertheless, the surprisal of a given word
can be used as a predictor of the reading time for that word, and
these predictions have been shown to correlate with reading measures
in eye-tracking corpora \citep{mcdonald_eye_2003,
  mcdonald_low-level_2003, demberg_data_2008, frank:bod:11,
  smith_effect_2013}.

The model proposed in this article has a surprisal component, which is
implemented using a neural language model. It departs from standard
surprisal by explicitly modeling which words in the input are attended
to, which enables it to predict word skipping. Skipping is a
particularly intriguing phenomenon in human reading: about 40\% of all
words in a text are skipped during reading (in the Dundee corpus, see
below), without apparent detriment to understanding. In a further
departure from standard surprisal, our model also models text
understanding explicitly, in the form of a task the model has to
perform. \changed{We experiment with two tasks: reconstructing the
  input text or answering questions about it. In both cases, the model
  needs a to learn a representation of the text that enables it to
  perform the task in an optimal way.} This optimization process is
what enables us to capture how reading behavior is affected by task
demands. All components of our model (surprisal, attention, task) are
implemented as neural networks and trained on unannotated text.

\subsection{Neural Network-based Attention}
\label{sec:neural_att_models}

Neural networks have been used for cognitive modeling since their
inception, with many of the fundamental architectures and algorithms
developed decades ago by \citet{PDP:86}. Neural models are attractive
in a cognitive context because they perform representation learning,
i.e., suitable representations for a given cognitive task emerge during
training, rather than having to be pre-specified by the model
designer. This provides a potential explanation for how cognitive
functions are acquired based on general learning mechanisms in
combination with the exposure to suitable training data.

The renaissance of neural networks in artificial intelligence was
triggered by the development of new learning algorithms, more advanced
network architectures, and the availability of massive amounts of data
and compute. It led to breakthroughs in areas such as computer vision
\citep{krizhevsky2012imagenet}, speech recognition
\citep{graves2013speech}, and natural language processing
\citep{collobert2011natural}. Today's neural architectures and
algorithms are efficient, scalable, and robust, enabling the
development of models that can be trained on large, realistic datasets
and achieve state-of-the-art performance on a wide range of AI tasks.

\changed{More recently, neural networks have been augmented with
  attention mechanisms. Attention in this context can be understood as
  a distribution over the input of the network that indicates which
  parts of the input are important for computing the network output.}
A range of attention-based neural network architectures have been
proposed in the literature, showing promise in both natural language
processing and computer vision \citep[e.g.,][]{mnih_recurrent_2014,
  bahdanau_neural_2015}.  These architectures either employ soft
attention or hard attention.  Soft attention distributes real-valued
attention values over the input and is typically trained using
backpropagation.  Hard attention mechanisms make discrete choices
about which parts of the input to focus on, and can be trained with
reinforcement learning \citep{mnih_recurrent_2014}. In language
processing models, soft attention can mitigate the difficulty of
compressing long sequences into fixed-dimensional vectors, with
applications in machine translation \citep{bahdanau_neural_2015} and
question answering \citep{hermann_teaching_2015}. In computer vision,
both types of attention can be used for selecting image regions for
processing \citep{ba_learning_2015, xu_show_2015}.

When it comes to modeling word skipping, i.e., the decision whether a
given word should be fixated or not, hard attention is a natural
modeling choice. It mimics the binary nature of the skipping decision
(rather than assuming a real-valued attention weight), and can be
trained using reinforcement learning, thus avoiding a cognitively
implausible supervised learning scheme.

%%%%%%%%%%%%%%%%%%%%%%%%%%%%%%%%%%%%%%%%%%%%%%%%%%%%%%%%%%%%%%%%%%%%%%%%%%%%%%%%
\section{The Tradeoff Hypothesis}
\label{sec:tradeoff}

Two observations about human reading are the starting point for our
model design: its efficiency and its task adaptivity.

Despite the complexity of the reading process, experienced readers are
able to process text very efficiently, at a typical rate of 200--400
words per minute \citep{Rayner:ea:16}. This is significantly higher
than the normal speaking rate of around 150 words per minute, meaning
that reading enables faster information uptake than listening
\changedSecond{at the normal rate.} Reading is a cultural skill,
subject to explicit teaching in school for a period of 10--14 years in
most countries. This means readers have a lot of practice by the time
they reach adulthood, and have honed their reading skills to the point
of automaticity; the Stroop effect \citep{Scarpina:17} is often cited
as evidence of such automaticity. Furthermore, the reading process is
efficient in that is able to deal with errors in the input. Even when
a text contains a high rate of letter transpositions and misspellings,
this does not seem to impair comprehension, and only marginally slows
down reading \citep{Hahn:ea:19}.

The second key observation is that humans adapt their reading strategy
to the task at hand. A considerable literature exists that
demonstrates how the reading task affects eye-movements, demonstrating
effects on, for instance, skipping rate, fixation duration, and
saccade length. Task is also known to interact with text-related
factors such as word length, word frequency, and word predictability.
\citet{Rayner:Fischer:96} compared normal reading with scanning
transformed text (every letter is replaces with a \emph{z}) and word
search. They found shorter fixations, longer saccades, and less
skipping in the normal reading condition. Also, word frequency effects
were absent in the scanning and search conditions.  Related to this,
\citet{Radach:ea:08} compared eye-movements in reading for
comprehension and in word verification, and found increased reading
times, as well as larger word frequency effect in comprehension.
\citet{Greenberg:ea:06} compared normal reading with a letter
detection task, and found longer fixations and less skipping for the
letter detection, but uncovered no task differences in word class and
predictability effects. In the same vein, \citet{White:ea:15} compared
reading for comprehension to the task of scanning for a specific
topic. In contrast to \citegen{Rayner:Fischer:96} word search results,
they found word frequency effects for both tasks in first fixation
times, but larger effects of word frequency in later measures during
reading for comprehension. When \citet{Kaakinen:Hyona:10} compared
reading for comprehension and proofreading for errors, they found that
the temporal and spatial properties of fixations and saccades differed
for the two tasks. They also found greater word length and word
frequency effects for proofreading compared to comprehension.
\citet{Schotter:ea:14} replicated these results and additionally
showed predictability effects in proofreading.  \citet{Kaakinen:ea:15}
compared the eye-movements of adults and children performing either a
question answering or a text comprehension task, and found task an
effects on first-pass reading in both groups; the adults also showed a
task effect on look-backs.

In summary, human reading is both very efficient and highly task
adaptable. We hypothesize that readers achieve these two remarkable
properties by optimizing a tradeoff between \emph{economy} and
\emph{accuracy}. Concretely, this means that they try to read as
efficiently as possible, for example by skipping as many words as they
can. At the same time, their reading behavior needs to ensure that
they make as few errors as possible in the task they are trying to
accomplish by reading.  We will call this assumption the
\emph{Tradeoff Hypothesis}. Based on this hypothesis, we expect that
humans only fixate words to the extent necessary for task success,
while skipping words that are not task relevant, or whose contribution
to the text can be inferred from context. Crucially, the optimal
economy--accuracy tradeoff depends on what the reader is trying to
accomplish; it follows that there is no single reading strategy that
is optimal everywhere; economy and accuracy trade off in different
ways for different tasks.

As mentioned in Section~\ref{sec:eye-movement_models}, the key
assumption that readers make a tradeoff is shared with
\citegen{bicknell_rational_2010-1} rational model of eye-movement
control in reading. Arguably, all Bayesian models involve a tradeoff,
as they assume that a prior probability distribution is combined with
a likelihood function in an optimal way. Another example is
\citegen{nor06} Bayesian model of word recognition, which trades off
recognition accuracy and recognition speed by combining a prior over
lexical properties (such as word frequency) with a likelihood function
that models noisy visual input.  \changed{Also, the model of
  \citet{Lewis:ea:13} assumes an explicit tradeoff between speed and
  accuracy to predict eye-movements in a list lexical decision
  task. They assume that information is integrated across saccades
  using a Bayesian update mechanism.}
  
In order to accrue evidence for the Tradeoff Hypothesis, this article
will investigate the following questions:
\begin{enumerate}

\item Can we use the Tradeoff Hypothesis to design a computational
  model that predicts quantitative properties of human skipping
  behavior? Furthermore, can we use this model to compute a surprisal
  measure that correlates with human reading time, even though it
  only has access to the words the model fixates (rather than to all
  words)?

\item Does this model instantiate a key prediction of the Tradeoff
  Hypothesis, viz., that the optimal reading strategy depends on the
  task? Based on this prediction, we expect that skipping behavior
  changes when the reading task changes, both in humans and in our
  model.

\end{enumerate}
To investigate these questions, we develop an architecture that
combines a language model based on recurrent neural networks with an
attention mechanism, two approaches that have been very successful in
natural language processing and in AI more generally. We train our
model on a large text corpus with an objective function that
implements the Tradeoff Hypothesis. We then evaluate the model's
reading behavior against human eye-tracking data, both from an
existing eye-tracking corpus and from a novel eye-tracking experiment
designed to study task effects.  Apart from the unlabeled text used as
training data and the model architecture, no further assumptions about
language structure need to be made -- in particular, no
\changed{explicit lexical or grammatical knowledge} is required, and
no eye-tracking data is used at training time. We show that
cognitively plausible behavior emerges as a result of devising an
architecture that can solve realistic reading tasks, and training it
on large amounts of unlabeled text.

%%%%%%%%%%%%%%%%%%%%%%%%%%%%%%%%%%%%%%%%%%%%%%%%%%%%%%%%%%%%%%%%%%%%%%%%%%%%%%%%
\section{The Neural Attention Tradeoff Model}
\label{sec:neat}

The point of departure for our model is the Tradeoff Hypothesis
introduced in the previous section: Reading optimizes a tradeoff
between economy of attention and task accuracy.  We make this idea
explicit by proposing NEAT (NEural Attention Tradeoff), a
computational model that reads text and then performs a task related
to the text it has read.  During reading, the network chooses which
words to process and which ones to skip. During normal reading, we
will assume that the task \changedSecond{the model needs to solve} is to \changed{to build up a representation
  of the text that allows it to reconstruct the input as
  accurately as possible.}  Based on this assumption, we train the
model using an objective function that minimizes the input
reconstruction error while also minimizing the number of words
fixated. We will drop this simplifying assumption and generalize our
model to a new reading task (question answering) in
Section~\ref{sec:neat_task}.

\changed{At this point, it is important to emphasize that our
  assumption
  \changedSecond{that text reconstruction is the default task that that our model performs} is clearly a simplification. We assume that during reading,
  the model learns a neural representation of the input text (stored
  in the weights of the network) that allows it to recall the
  input. We achieve this by training the model to reconstruct the
  input text as accurately as possible from its neural
  representation. However, by making this modeling assumption, we do
  not mean to imply that human readers memorize what they read
  verbatim; rather, it seems plausible that they construct an
  integrated representation of the propositions expressed in the text,
  as well as a mental model of the situation described, as assumed by
  \citet{kintsch1988role}. A recall of the information in the text
  would then be achieved by querying these integrated
  representations. However, given that it is not currently feasible to
  automatically build up Kintsch-type text representations for
  naturalistic text (such as the newspaper text we use), we make the
  simplifying assumption that the model learns a representation that
  can be used for input reconstruction.}

To reiterate, the aim of this article is not to offer a full model of
eye-movement control during reading. Rather, we want to model one
particular aspect of reading, i.e., the allocation of attention to the
words in a text. Specifically, we want to capture the process that
decides which words in a text are fixated and which ones are skipped
when a text is read. Once we have a model of skipping, we can use that
model to compute a modified version of the surprisal that only uses
the words that are fixated. This in turn allows us to compute
word-by-word reading times.

\subsection{Architecture}
\label{sec:architecture}

Our architecture is based on the standard neural encoder-decoder
architecture, which encodes a read sequence into memory
representations and then decodes output from it
\citep{kalchbrenner2013recurrent,sutskever_sequence_2014,cho2014learning}.
We illustrate the model in Figure~\ref{fig:architecture}, operating on
a three-word sequence $\boldsymbol{w} = w_1, w_2, w_3$.  The basic
components of NEAT are the \emph{reader}, labeled $R$, and the
\emph{decoder}.  Both these components are recurrent neural networks
with long short-term memory (LSTM, \citealp{hochreiter_long_1997})
units.  The recurrent reader network is expanded into time steps $R_0,
\dots, R_3$ in the figure.  It passes over the input sequence, reading
one word $w_i$ at a time, and converts the word sequence into a
sequence of vectors $h_0, \dots, h_3$.  Each vector $h_i$ is as a
fixed-dimensional encoding of the word sequence $w_1, \dots, w_i$ that
has been read so far.  The last vector $h_3$ (more generally $h_N$ for
a sequence of length $N$), which encodes the entire input sequence, is
then fed into the input layer of the decoder network, which attempts
to reconstruct the input sequence $\boldsymbol{w}$. The decoder is
also realized as a recurrent neural network, collapsed into a single
box in the figure.  It models a probability distribution over word
sequences $w'$, computing the conditional distribution
$P_{Decoder}(w'_i|\boldsymbol{w}'_{1 \dots i-1}, h_N)$ over the
vocabulary in the $i$-th step, as is common in neural language
modeling \citep{mikolov_recurrent_2010}. As the decoder has access to
the vector representation created by the reader network, it should be
able to assign the highest probability to the word sequence
$\boldsymbol{w}$ that was actually read.  Up to this point, the model
is a standard encoder-decoder architecture designed to reconstruct an
input sequence from a fixed-dimensional representation.

Secondly, we model skipping by stipulating that only some of the input
words $w_i$ are fed into the reader network $R$; the other words are
skipped during reading. For skipped words, $R$ receives a special
vector representation that contains no information about the input
word.  NEAT incorporates an \emph{attention module} $A$, which at each
time step during reading, decides whether the next word is shown to
the reader network or not. Before fixating or skipping a word, humans
obtain information about it and sometimes fully identify it using
\emph{parafoveal preview} \citep{gordon_see_2013}.  Thus, we can
assume that the choice of which words to skip takes into account a
preview of the word itself. \changed{We model parafoveal preview by
  assuming that the attention module receives the first four
  characters of the input word when making its decision.}

Note that the literature provides mixed evidence regarding the role of
context in shaping skipping decisions.  There is evidence for strong
frequency effects that trump effects of contextual fit
\citep{angele2013processing, angele2014the}, but studies have also
found evidence for contextual predictability effects
\citep{balota1985the, kliegl2004length, rayner2011eye, luke2016limits,
  duan2020rational}.  To test whether contextual information is
relevant for predicting skipping with our model, we consider two
versions of NEAT: In the \emph{NoContext} version, skipping is based
only on the identity of the word.  In the \emph{WithContext} version,
the attention module has additionally access to the previous state
$h_{i-1}$ of the reader network, which summarizes what has been read
so far. This is represented by a dashed line in
Figure~\ref{fig:architecture}.

If we write the decision made by $A$ as $\omega_i \in \{0, 1\}$, where
$\omega_i = 1$ means that word $w_i$ is shown to the reader and $0$
means that it is skipped, then we can write the probability of showing
word $w_i$~as:
\begin{equation}\label{eq:att-prob}
  P(\omega_{i}=1|\boldsymbol{\omega}_{1 \dots i-1}, \boldsymbol{w})
	= A(w_i, h_{i-1} )  % P_{R}(w_i|\boldsymbol{w}_{1 \dots i-1},  \boldsymbol{\omega}_{1 \dots i-1})
\end{equation}
\changed{where $A(w_i, h_{i-1})$ is the output of the attention module $A$.}
We implement the attention module $A$ as a feed-forward network,
followed by taking a binary sample~$\omega_i$.

As outlined in Section~\ref{sec:processing_diff_models},
\emph{surprisal} is a prominent model of linguistic processing
difficulty that has been used to capture human reading time. Standard
surprisal models assume that all words are read, and it is an
interesting question whether surprisal estimates computed on noisy word
sequences (where some words have been skipped) are able to predict
reading time correctly.

The surprisal of an input word $w_i$ is the negative logarithm of the
conditional probability of the word given the context words
$\boldsymbol{w}_{1 \dots i-1}$. 
\changedSecond{We estimate surprisal using a second LSTM network that at each time step computes a probability distribution $P_S$ over the lexicon, and is trained to predict the next word given the fixated words. Using the skipping sequence $\boldsymbol{\omega}_{1 \dots i-1}$ predicted by NEAT, surprisal becomes:}
\begin{equation}\label{eq:surp}
  \operatorname{Surp}(w_i|\boldsymbol{w}_{1 \dots i-1}) = -\log
  P_{S}(w_i|\boldsymbol{w}_{1 \dots i-1}, \boldsymbol{\omega}_{1 \dots i-1})
\end{equation}
Crucially, this surprisal estimate only takes into account the words
that have actually been read. We will refer to this quantity as
\emph{restricted surprisal}, as opposed to \emph{full surprisal},
which is computed based on all prior context words, and \changed{constitutes a best possible estimate of the surprisal.}

NEAT therefore computes two quantities that model important aspects of
reading behavior: the fixation probability in
equation~(\ref{eq:att-prob}), which predicts how likely words are to
be fixated (rather than skipped), and the restricted surprisal in
equation~(\ref{eq:surp}), which models the reading times of fixated
words (in line with standard results that show that surprisal
correlates with reading time, e.g.,~\citealp{demberg_data_2008}).

\def\la{-7.7}
\def\lb{-4.0}
\def\lc{-0.3}
\def\ld{3.4}
\def\le{5.3}
\def\laa{-5.5}
\def\lba{-2.0}
\def\lca{1.7}

\begin{figure}[]
\centering
\begin{tikzpicture}[%
  % common options for blocks:
  block/.style = {draw, fill=blue!30, align=center, anchor=west,
              minimum height=0.65cm, inner sep=0},
  % common options for the circles:
  ball/.style = {circle, draw, align=center, anchor=north, inner sep=0}]

%\draw[domain=-3.5:0.5] plot (\x,{-3.5});
%\draw[domain=-3.5:0.5] plot (\x,{1});
%\draw[domain=-3.5:1] plot ({-3.5},\x);
%\draw[domain=-3.5:1] plot ({0.5},\x);

\node[rectangle,draw=none,text width=0.3cm,anchor=base] (W1) at (\laa,-1) {$w_1$};
\node[rectangle,draw=none,text width=0.3cm,anchor=base] (W2) at (\lba,-1) {$w_2$};
\node[rectangle,draw=none,text width=0.3cm,anchor=base] (W3) at (\lca,-1) {$w_3$};

\node[rectangle,draw,text width=0.3cm,anchor=base] (A1) at (\lb,-1) {A};
\node[rectangle,draw,text width=0.3cm,anchor=base] (A2) at (\lc,-1) {A};
\node[rectangle,draw,text width=0.3cm,anchor=base] (A3) at (\ld,-1) {A};

\node[rectangle,draw,text width=0.3cm,anchor=base] (R0) at (\la,-3) {$R_0$};
\node[rectangle,draw,text width=0.3cm,anchor=base] (R1) at (\lb,-3) {$R_1$};
\node[rectangle,draw,text width=0.3cm,anchor=base] (R2) at (\lc,-3) {$R_2$};
\node[rectangle,draw,text width=0.3cm,anchor=base] (R3) at (\ld,-3) {$R_3$};

\node[rectangle,draw,anchor=base] (Task) at (\le,-3) {Decoder};

\draw[->] (R0.east) to [out=0,in=180] (R1.west);
\draw[->] (R1.east) to [out=0,in=180] (R2.west);
\draw[->] (R2.east) to [out=0,in=180] (R3.west);
\draw[->] (R3.east) to [out=0,in=180] (Task.182);

% The following three dashed lines are connections from R to A
\draw[dashed, ->] (R0.east) to [out=10,in=250] node[pos=0.65]{$\ \ \ \ \ \ \ h_0$} (A1.225);
\draw[dashed, ->] (R1.east) to [out=10,in=250]  node[pos=0.65]{$\ \ \ \ \ \ \ h_1$} (A2.225);
\draw[dashed, ->] (R2.east) to [out=10,in=250]  node[pos=0.65]{$\ \ \ \ \ \ \ h_2$} (A3.225);

\draw[->] (W1.10) to [out=0,in=180] (A1.west);
\draw[->] (W2.10) to [out=0,in=180] (A2.west);
\draw[->] (W3.10) to [out=0,in=180] (A3.west);

\draw[->] (A1.south) to [out=270,in=90] (R1.north);
\draw[->] (A2.south) to [out=270,in=90] (R2.north);
\draw[->] (A3.south) to [out=270,in=90] (R3.north);

\end{tikzpicture}

\hspace{1ex}

\caption{The architecture of the proposed NEAT model, reading a
  three-word input sequence $w_1, w_2, w_3$.  $R$ is the reader
  network. $A$ is the attention network. At each time step, the input
  $w_i$ is fed into $A$, which then decides whether the word is read
  or skipped. In the \emph{WithContext} version, the attention module
  additionally has access to the state $h_{t-1}$ of the reader
  network, indicated by dashed lines.}
  \label{fig:architecture}
\end{figure}

\subsection{Objective Function}
\label{sec:objective}

Given the network parameters $\theta$ and an input sequence of words
$\boldsymbol{w}$, the network stochastically chooses a \changed{vector
  $\boldsymbol{\omega}$ of fixation decisions} 
according to~(\ref{eq:att-prob}) and
incurs a loss $L(\boldsymbol{\omega}|\boldsymbol{w},\theta)$, \changedSecond{which we take to be the negative log-probability assigned by the decoder to the correct input sequence:}
%when
%reconstructing the input sequence: % language modeling (i.e.,~word
                                   % prediction) and
%
\begin{eqnarray}
	L(\boldsymbol{\omega}|\boldsymbol{w}, \theta) &=& - \frac{1}{N} \sum_{i=1}^N \log P_\mathit{Decoder}(w_i|\boldsymbol{w}_{1 \dots
     i-1};  h_N; \theta) %\nonumber
\end{eqnarray}
where
%$P_R(w_i,\dots)$ denotes the output of the reader after reading
%$w_{i-1}$, and
$P_\mathit{Decoder}(w_i|\boldsymbol{w}_{1 \dots i-1}; h_N; \theta)$ is
the probability assigned by the decoder network to the $i$-th
  input word \changedSecond{assuming it has correctly generated the preceding words}, with $h_N$ being the vector representation created by
the reader network for the entire input sequence.

To implement the Tradeoff Hypothesis, we train NEAT to solve
reconstruction with a minimal number of fixations, i.e., the network minimizes the
following expected loss:
\begin{equation}\label{eq:objective}
  Q(\theta) := \operatorname{E}_{\boldsymbol{w},
    \boldsymbol{\omega}}\left[L(\boldsymbol{\omega}|\boldsymbol{w},
	\theta) + \alpha \cdot \frac{\|\boldsymbol{\omega}\|_{\ell_1}}{N}\right]
\end{equation}
where word sequences $\boldsymbol{w}$ are drawn from a corpus, and
$\boldsymbol{\omega}$ is distributed according to
$P(\boldsymbol{\omega}|\boldsymbol{w}, \theta)$ as defined
in~(\ref{eq:att-prob}).  In equation~(\ref{eq:objective}),
$\|\boldsymbol{\omega}\|_{\ell_1}$ is the number of words shown to the
reader, and $\alpha > 0$ is a hyperparameter, i.e., a parameter that
needs to be tuned on a development set.  The term $\alpha \cdot
\|\boldsymbol{\omega}\|_{\ell_1}$ encourages NEAT to attend to as few
words as possible; the bigger $\alpha$ is, the higher the loss
incurred when reading (as opposed to skipping) words.

Note that we make no assumption about linguistic structure -- the only
ingredients of NEAT are the neural architecture, the
objective~(\ref{eq:objective}), and the corpus from which the
sequences $\boldsymbol{w}$ are drawn. This corpus is merely a sequence
of words -- NEAT is trained without linguistic annotation and without
any eye-tracking data; we use eye-tracking data only for evaluating
the model.

\subsection{Parameter Estimation}
\label{sec:parameters}

We jointly train the parameters of the reader, decoder, and attention
modules to minimize the objective function~(\ref{eq:objective}).  We
use gradient descent for the reader and decoder modules, and
reinforcement learning, specifically the REINFORCE method
\citep{williams_simple_1992}, for the attention module~$A$.

Starting with random initial values, the parameters are updated
iteratively as follows.  In each iteration, a text is selected from
the training corpus, and a sequence of fixations and skips
$\boldsymbol{\omega} = \omega_1, \dots, \omega_N$ is sampled using the
NEAT attention module $A$ given the current parameter setting
$\theta$.  We then compute the gradients as follows:
\begin{eqnarray}\label{eq:parameter-update}
	\Delta_1 &:=&  \left(\nabla_\theta \log \prod_{i=1}^N P_A(\omega_i|\omega_{1\dots i-1}
	{\bf w}; \theta)\right) \cdot
\left[L(\boldsymbol{\omega}|\boldsymbol{w},
    \theta) + \alpha \cdot \|\boldsymbol{\omega}\|_{\ell_1}\right] \\
\Delta_2 &:=&  \nabla_\theta L(\boldsymbol{\omega}|\boldsymbol{w},    \theta)	\nonumber
\end{eqnarray}
\changed{where $\nabla_\theta$ denotes the gradient with respect to the parameters $\theta$,}
using the backpropagation algorithm and update the parameters $\theta$
according to the following update rule:
\begin{equation}\label{eq:update-rule}
\theta \leftarrow \theta - \gamma\cdot(\Delta_1+\Delta_2)
\end{equation}
where $\gamma > 0$ is the learning rate.  To speed up computation, in
every iteration, 64 texts are batched together, and all computations
are performed in parallel on them.

This technique can be understood as a reinforcement learning method
where the attention module samples actions from its current strategy
and updates its parameters based on a reward that trades off accuracy
and attention.  The update rule in equation~(\ref{eq:update-rule}) is
also an instance of stochastic gradient descent, since $\Delta_1 +
\Delta_2$ is an unbiased estimator of the gradient
of~(\ref{eq:objective}) with respect to~$\theta$
\citep{williams_simple_1992}. \changedSecond{This method is widely used in the reinforcement learning literature, and commonly referred to as the policy-gradient method.}

We utilize three methods for speeding up convergence of this training
algorithm.  First, when computing the
gradient~(\ref{eq:parameter-update}), we subtract from the reward
signal $L(\boldsymbol{\omega}|\boldsymbol{w}, \theta) + \alpha \cdot
\|\boldsymbol{\omega}\|_{\ell_1}$ a running average of its recent
values on other texts, a standard approach for speeding up
reinforcement learning without sacrificing the unbiasedness of the
gradients~\citep{williams_simple_1992}.  From a cognitive perspective,
this corresponds to adjusting the reward signal based on the value of
the loss \emph{relative to typical values} on other texts.  Second,
following \cite{xu_show_2015}, we add entropy regularization,
encouraging the model to explore different reading strategies.  Third,
we pretrain the reading and decoder networks on input sequences where
5\% of words have been skipped at random. This corresponds to
collecting prior knowledge about the language before starting to
optimize the reading strategy. % ($\boldsymbol{\omega} \sim
%\mathrm{Binom}(n,p)$ ($n$ sequence length, $p$ a hyperparameter).

%%%%%%%%%%%%%%%%%%%%%%%%%%%%%%%%%%%%%%%%%%%%%%%%%%%%%%%%%%%%%%%%%%%%%%%%%%%%%%%%
\section{Modeling Study~1}
\label{sec:study1}

In this study, we will evaluate our NEAT model against human
eye-tracking data. In particular, we want to investigate whether NEAT,
when trained to perform a generic task (reconstructing the input
text), is able to simulate reading behavior as it is observed in human
readers when they are not given a specific task (other than
understanding the text).

NEAT predicts two quantities that can be evaluated against human
reading data: \changed{Firstly, it generates vectors of skips and
  fixations, i.e., which words in a given input text are fixated or
  skipped.}  Secondly, it predicts surprisal values for words in
context. We can evaluate model generated fixation sequences against
human fixation \changed{vectors}, and surprisal values against human
reading time. Both can be extracted from eye-tracking corpora.

\subsection{Methods}
\label{sec:study1:methods}

\subsubsection{Model Implementation}\label{sec:study1:implementation}

For both the reader and the decoder networks, we choose a one-layer
LSTM network with 1,024 memory cells.  The attention network applies a
linear transformation to word embeddings and (in the
\emph{WithContext} version) the previous reader state, and derives a
probability using the logistic function:
\begin{align}\label{eq:attention-study1}
  P(\omega_{i}=1|\boldsymbol{\omega}_{1 \dots i-1}, \boldsymbol{w}) &:= \sigma\left(u + v^T {\hat w}_i\right) \ \ \ \ \ \ \ \ \ \ \ \ \ \ \ \ \ \ \ \ \text{{NoContext}} \\
  P(\omega_{i}=1|\boldsymbol{\omega}_{1 \dots i-1}, \boldsymbol{w}) &:= \sigma\left(u + v^T {\hat w}_i + h_{i-1}^T A {\hat w}_i\right) \ \ \ \text{{WithContext}}
\end{align}
where $\hat w_i \in \mathbb{R}^{d_{emb}}$ is a \changed{vector encoding the preview of the word $w_i$ (see below)}.  The weights $v \in \mathbb{R}^{d_{emb}}$, $A \in \mathbb{R}^{d_{emb}
  \times 1024}$ are parameters of the model.
\changed{We encoded the vectors for the preview as follows. First, for each prefix of four characters from the model's vocabulary, we allocated 200-dimensional embedding vector that is initialized as the average of the word embeddings of all words continuing this prefix. These embedding vectors were then trained together with the other parameters of the attention network. Second, in order to make character-level information directly accessible, we represent each character in the preview prefix with a small (128-dimensional) vector trained together with the other parameters, average these, and concatenate the resulting vector with the 200-dimensional vector for the prefix, leading to an overall representation with $d_{emb}=200+128$ dimensions.}

The text being read is split into sequences of 50 tokens, which are
used as the input sequences for NEAT, disregarding sentence
boundaries.  Word embeddings are shared between the reader and the
attention network.  The vocabulary consists of the 50,000 most
frequent words from the training corpus. \changedSecond{Words outside of this vocabulary were represented with a special out-of-vocabulary token, as is standard practice in neural sequence modeling of language; this affected 2.0\% of tokens in the training set.} We trained NEAT on the
\changed{training set of the DeepMind question answering corpus}
\citep{hermann_teaching_2015}, which consists of 90,266 articles the
from CNN web site \changed{and 196,961 articles from the Daily Mail
  web site, containing approximately 230 million tokens} (only the
texts were used for this modeling study, not the questions).
\changed{During training, texts were sampled from both sources in
  equal proportion.}  Training iterated through the corpus four
times. For initialization, weights are drawn from the uniform
distribution.

In order to determine the weight $\alpha$ in (\ref{eq:objective}), we
varied $\alpha$ from $0$ to $3$ in steps of $0.25$, and chose the
value ($\alpha = 3.0$) resulting in the fixation rate (63.5\%) that
best matched the human
fixation rate on the development set (see Section~\ref{sec:study1:dataset}).
%We used a constant learning rate for~$A$, varied among 0.1, 0.2, 0.5, 1 across model runs.
No eye-tracking data was used when training the model, beyond the choice of
$\alpha$ to match human fixation rate.

\subsubsection{Dataset}
\label{sec:study1:dataset}

To evaluate the reading behavior of the trained model, we used the
English section of the Dundee corpus
\citep{kennedy_parafoveal--foveal_2005}, which consists of 20~texts
from \emph{The Independent}, annotated with eye-movement data from ten
English native speakers. Each native speakers read all 20~texts and
answered a comprehension question after each text.  We split the
Dundee corpus into a development and a test set, with texts 1--3
constituting the development set.  The development set consists of
78,300 tokens, and the test set of 281,911 tokens.  For evaluation, we
removed the datapoints removed by \cite{demberg_data_2008}, mainly
consisting of words at the beginning or end of lines and
cases of track loss.  Furthermore, we removed datapoints where the
word was outside of the vocabulary of the model.  After preprocessing,
81.1\% of tokens remained. To obtain fixation rate and reading time
per token, we used the eye-tracking measures computed by
\cite{demberg_data_2008}.  The average fixation rate over all tokens
was 62.1\% on the development set, and 61.3\% on the test set.

The development set was used to determine the human skipping rate (see
Section~\ref{sec:study1:results}) and to choose the model
parameter $\alpha$ (Section~\ref{sec:study1:implementation}).
The model was then run on the test set to compute the
results that we report in the following section.

\subsection{Results and Discussion}
\label{sec:study1:results}

\subsubsection{Fixation Vectors}
\label{sec:study1:results:fix}

\begin{figure}[]

%\definecolor{color0}{rgb}{0,0.2,1}
%\definecolor{color1}{rgb}{0.1,0.2,1}
%\definecolor{color2}{rgb}{0.2,0.2,1}
%\definecolor{color3}{rgb}{0.3,0.2,1}
%\definecolor{color4}{rgb}{0.4,0.2,1}
%\definecolor{color5}{rgb}{0.5,0.2,1}
%\definecolor{color6}{rgb}{0.6,0.2,1}
%\definecolor{color7}{rgb}{0.7,0.2,1}
%\definecolor{color8}{rgb}{0.8,0.2,1}
%\definecolor{color9}{rgb}{0.9,0.2,1}
%\definecolor{color10}{rgb}{1,0.2,1}
%\definecolor{color11}{rgb}{1,0.2,0.9}
%\definecolor{color12}{rgb}{1,0.2,0.8}
%\definecolor{color13}{rgb}{1,0.2,0.7}
%\definecolor{color14}{rgb}{1,0.2,0.6}
%\definecolor{color15}{rgb}{1,0.2,0.5}
%\definecolor{color16}{rgb}{1,0.2,0.4}
%\definecolor{color17}{rgb}{1,0.2,0.3}
%\definecolor{color18}{rgb}{1,0.2,0.2}
%\definecolor{color19}{rgb}{1,0.2,0.1}
%\definecolor{color20}{rgb}{1,0.2,0}
%\definecolor{colorNaN}{rgb}{0.99,0.99,0.99}

\centering

Human
	
	\vspace{2ex}

	\includegraphics[width=.9\textwidth]{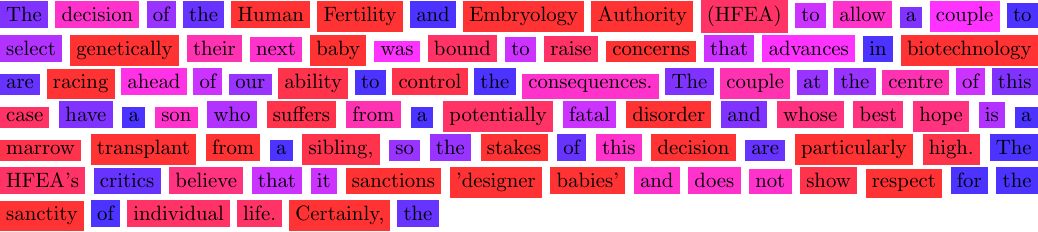}

\vspace{2ex}
Model
\vspace{2ex}

\includegraphics[width=.9\textwidth]{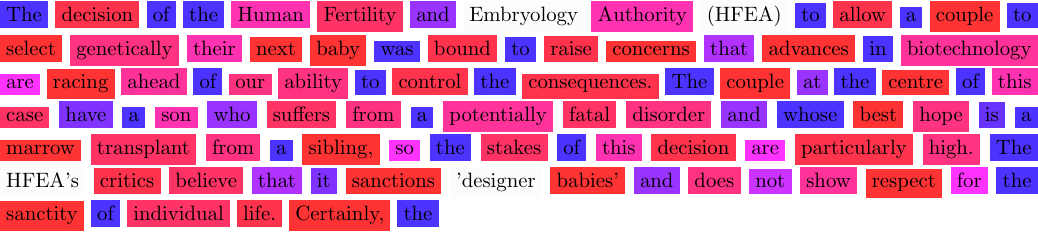}

\vspace{2ex}

\caption{Top: Heatmap visualizing human fixation probabilities, as
  estimated from the ten readers in the Dundee corpus. In cases of
  track loss, we replaced the missing value with the corresponding
  reader's overall fixation rate. Bottom: Heatmap showing fixation
  probabilities simulated by NEAT in Modeling Study~1. The color
  gradient denotes the fixation probability for a word, ranging from
  blue (low probability) to red (high probability). White background
  indicates that the word was excluded from the analysis because it
  was not part of the model's vocabulary.
	}
\label{fig:heatmaps}
\end{figure}

\begin{table}[]
\begin{center}
\begin{tabular}{llll}
\hline
& Accuracy & $F$$_{\mathrm{fix}}$ & $F$$_{\mathrm{skip}}$ \\ \hline
NEAT (NoContext) & 66.9 & 71.2 & 61.0 \\ 
NEAT (WithContext) & 66.6 & 71.0 & 60.6\\
\hline
\multicolumn{4}{c}{Supervised Models} \\ \hline
\cite{nilsson_learning_2009} & 69.5 & 75.2 & 62.6 \\
\cite{matthies_blinkers_2013} &  69.9 & 72.3 & 66.1 \\ \hline
\multicolumn{4}{c}{Human Performance and Baselines} \\ \hline
Random Attention & 51.2 & 58.1 & 41.4 \\
Full Surprisal & 61.5 & 66.6 & 54.6 \\
Word Frequency & 67.4 & 71.7 & 61.7 \\
Word Length & 69.1 & 72.7 & 64.5 \\
Human Agreement & 69.3 & 73.2 & 64.1 \\
\hline
\end{tabular}
\end{center}
\caption{Evaluation of fixation predictions against human
  data. To measure human agreement, we predicted the $n$-th reader's
  fixations by taking the fixations of the $(n+1)$-th reader (with
  missing values replaced by reader average), averaging the resulting
  scores over the ten readers.}\label{tab:disc-eval}
\end{table}

NEAT defines for each word in a sequence a probability that it will be
fixated.  These simulated fixation probabilities can be interpreted as
defining a distribution of attention over the input sequence.
Figure~\ref{fig:heatmaps} shows heatmaps of the human and simulated
fixation probabilities (\emph{NoContext} version), respectively, for
the beginning of a text from the Dundee corpus.  While some
differences between human and simulated fixation probabilities can be
noticed, there are similarities in the general qualitative features of
the two heatmaps.  In particular, function words and short words are
less likely to be fixated than content words and longer words in both
the human and the simulated data.

%%%%%%%%%%%%%%%%
%%%%%%%%%%%%%%%%
%%%%%%%%%%%%%%%%
%%%%%%%%%%%%%%%%

\changed{In order to quantitatively assess fit to human fixation
  rate, we compared NEAT fixation probabilities to the following
  models.  First, we compare to previous work on supervised models of
  fixation prediction on the Dundee corpus
  \citep{nilsson_learning_2009,matthies_blinkers_2013}.  Second, we
  considered \emph{random attention}, which is defined by
  $\boldsymbol{\omega} \sim \mathrm{Binom}(n,p)$, with $p = 0.621$,
  the human fixation rate in the development set. This baseline
  corresponds to a reading strategy where words are skipped at random,
  so that the overall fixation rate is the same as in the development
  portion of the Dundee corpus.  Third, we derived fixation
  predictions from \emph{full surprisal}, \emph{word frequency}, and
  \emph{word length}. For this, we choose a threshold for these
  quantities such that the resulting fixation rate matches the human
  fixation rate on the development set. For example, the word
  frequency baseline skips all high frequency words up to a frequency
  threshold that ensures that the fixation rate is 62.1\%. Note that
  we need a surprisal model in order to compute the full surprisal
  baseline; we will return to this below.  To enable meaningful
  comparison, we applied the same procedure to the NEAT fixation
  probabilities.\footnote{\changed{Another possibility would be to
      directly evaluate fixations \emph{sampled} from NEAT, but the
      results would not be comparable with the other models
      \citep{nilsson_learning_2009,matthies_blinkers_2013} -- where
      reported results refer to their \emph{highest probability}
      fixation sequences -- and to the baselines, for which fixation
      rate is matched to the human data.}}}

%%%%%%%%%%%%%%%%
%%%%%%%%%%%%%%%%
%%%%%%%%%%%%%%%%
%%%%%%%%%%%%%%%%

As in previous work, we report the classification accuracy (for the
two classes fixation and skip), and also separate $F$-scores for
fixation and skip prediction ($F$ is the harmonic mean of precision
and recall). 
\changedSecond{We computed a separate accuracy for each human subject and averaged the results, following \citet{matthies_blinkers_2013}.}
As lower and upper bounds, we use random attention
(defined as $\boldsymbol{\omega} \sim \mathrm{Binom}(n, 0.621)$) and
the agreement of the ten human readers, respectively.  The results are
shown in Table~\ref{tab:disc-eval}.  NEAT clearly outperforms random
attention and shows results close to full surprisal (where we apply
the same rescaling and thresholding as for NEAT).  This is remarkable
given that NEAT has access to only 63.5\% of the words in the corpus
in order to predict skipping, while full surprisal has access to all
the words.

When comparing the two different versions of NEAT, we find that access
to the previous context of a word does not result in improved fit; the
\emph{NoContext} model performs slightly better than the
\emph{WithContext} one.

Word frequency and word length perform well, almost reaching the
performance of supervised models.  This shows that the bulk of
skipping behavior is already explained by word frequency and word
length effects.  \changed{Note, however, that NEAT is completely
  unsupervised, and does not know that it has to pay attention to word
  frequency or word length; this is something the model is needs to
  infer.  A simple regression based on word frequency and word length
  would have fewer parameters than a neural network model, but it is
  not explanatory: it does not provide a theory of why those
  predictors should matter. In contrast, in NEAT, effects of these
  predictors are emergent properties of the Tradeoff Hypothesis.}

The success of word frequency and word length is in agreement with the
finding that contextual information did not improve model fit, and
with prior experimental evidence indicating that frequency effects can
trump contextual fit \citep{angele2013processing, angele2014the}.
While some studies have found evidence for contextual predictability
effects in skipping \citep{balota1985the, kliegl2004length,
  rayner2011eye, luke2016limits, duan2020rational}\changedSecond{,
  other work has found their role to be limited
  \citep{Heilbron2021PredictionAP}}.  Our results suggest that context
effects, as they are modeled by the \emph{WithContext} version of NEAT,
play no role for skipping in naturalistic text.

\subsubsection{Reading Time}
\label{sec:study1:results:rt}

To evaluate the predictions NEAT makes for reading time, we use
linear mixed effects models \citep{Pinheiro:Bates:00} that include as
a predictor the restricted form of surprisal derived from NEAT for the
Dundee test set. The mixed models also include a set of standard
baseline predictors, viz., word length, log word frequency, and the
position of the word in the text.  Word length and surprisal were
residualized with respect to log word frequency.  To keep the size of
the mixed effects models manageable, we only considered binary
interactions. Which interactions to include was determined by forward
model selection: Starting from a model with only main effects, we
iteratively added the binary interaction resulting in the greatest
improvement in deviance,\footnote{\changedSecond{Deviance is defined as the
  log likelihood of the model under
  consideration,  multiplied by $-2$.}} until model fit did not change significantly
any more according to a $\chi^2$ test.  We did this separately for all
four reading measures, and then pooled the interactions, so that the
final models for all measures contained the same set of interactions.

We treat participants and items as random factors.  As the dependent
variables, we take either first fixation duration, first pass time,
total time, or fixation rate (these measures are defined in
Section~\ref{sec:data_analysis} below).

We compare NEAT surprisal against \emph{full surprisal} as an upper
bound and against \emph{random surprisal} as a lower bound. Random
surprisal is surprisal computed by a model with random attention; this
allows us to assess how much surprisal degrades when only 63.5\% of
all words are fixated, but no information is available as to which
words should be fixated. Full surprisal is the surprisal estimate
computed with full attention, i.e., when our model is allowed to
fixate all words.

In order to evaluate the different surprisal estimates, \changedSecond{we compare the model fit} of the three mixed effects models using deviance:
higher deviance indicates greater improvement in model fit over the baseline model.
For first pass, we
find that the mixed model that includes NEAT surprisal \changedSecond{reduces deviance by}~\changed{413} compared to the mixed model containing only the
baseline predictors. With full surprisal, we obtain a \changedSecond{deviance difference}
of~\changed{499}. On the other hand, the model including random surprisal
achieves a lower \changedSecond{deviance difference} of~\changed{304}.  For the other reading time
measures, the situation is similar, see
Table~\ref{tab:neat-predicts-dundee} for details. (This table also
contains AIC and BIC as additional measures for model comparison;
these agree with the deviance results.)  For fixation rate, we find
that no form of surprisal has any predictive power over and above the
other predictors, in line with what we found for skipping.

The reading time results in Table~\ref{tab:neat-predicts-dundee} show
that restricted surprisal as computed by NEAT not only significantly
predicts reading time, it also provides an improvement in model fit
compared to the baseline predictors. The magnitude of this improvement
in terms of deviance indicates that NEAT outperforms random
surprisal. Full surprisal achieves an even greater improvement, but
this is not unexpected, as full surprisal has access to all words,
unlike NEAT or random surprisal, which only read to 63.5\% of the
words in the text, and skip the rest.\footnote{\changed{Full surprisal
    predicts reading time better than restricted surprisal. We could
    hypothesize that this indicates that words that are not fixated
    are in fact processed in some way, rather than truly
    skipped. However, this would only be true if all human readers in
    the Dundee corpus skipped the same words. In reality, individual
    participants will each skip a certain proportion of words, but
    they don't all skip the same words. Full surprisal was trained
    using a model that doesn't skip any words, so it can provide the
    best possible prediction of reading time for every participant, no
    matter which words they skip. Restricted surprisal, on the other
    hand, is trained on a model that skips words (but not necessarily
    the ones skipped by any one participant), and thus provides a
    worse estimate of reading time for an individual participant.}}

\begin{table}[]
\begin{center}
%\begin{tabular}{l@{}d@{}d@{~~}l@{~~}d@{}d@{~~}l@{~~}d@{}d@{~~}l@{~~}d@{}d@{~~~~}l@{}}
%\begin{tabular}{l@{}r@{}r@{~~}r@{}r@{~~}r@{}r@{~~}@{}r@{~~~~}}
\begin{tabular}{lrrrrrrr}
  \hline
& \mc{2}{c}{First Fixation}& \mc{2}{c}{First Pass} & \mc{2}{c}{Total Time}  \\ \hline
(Intercept) & \textbf{200.84} & (5.73) & \textbf{232.71} & (7.84) & \textbf{243.36} & (8.34) \\
WordLength & \textbf{3.54} & (0.23) & \textbf{32.82} & (0.43) & \textbf{35.54} & (0.48) \\
LogWordFreq & \textbf{-4.66} & (0.23) & \textbf{-2.99} & (0.41) & \textbf{-4.92} & (0.45) \\
PositionText & \textbf{-0.65} & (0.18) & \textbf{-2.3} & (0.32) & \textbf{-2.91} & (0.35) \\
LogWordFreq:WordLength & \textbf{0.37} & (0.17) & -0.62 & (0.32) & \textbf{-1.52} & (0.35) \\
LogWordFreq:PositionText & \textbf{-0.54} & (0.22) & \textbf{-0.96} & (0.40) & -0.31 & (0.45) \\
WordLength:PositionText & \textbf{-0.78} & (0.22) & \textbf{-2.94} & (0.41) & \textbf{-2.74} & (0.46) \\
\hline NEAT Surprisal  & \textbf{1.54} & (0.08) & \textbf{3.48} & (0.14) & \textbf{4.21} & (0.15) \\
\quad AIC Difference & 411.0 &  & 633.0 &  & 749.0 &  \\
\quad BIC Difference & 400.0 &  & 623.0 &  & 738.0 &  \\
	\quad \changedSecond{Deviance Difference} & 413.0 &  & 635.0 &  & 751.0 &  \\
\hline Full Surprisal  & \textbf{1.45} & (0.06) & \textbf{3.2} & (0.12) & \textbf{4.03} & (0.13) \\
\quad AIC Difference & 497.0 &  & 735.0 &  & 947.0 &  \\
\quad BIC Difference & 486.0 &  & 725.0 &  & 937.0 &  \\
\quad \changedSecond{Deviance Difference} & 499.0 &  & 737.0 &  & 949.0 &  \\
\hline Random Surprisal  & \textbf{1.21} & (0.07) & \textbf{2.65} & (0.13) & \textbf{3.18} & (0.14) \\
\quad AIC Difference & 302.0 &  & 439.0 &  & 513.0 &  \\
\quad BIC Difference & 292.0 &  & 429.0 &  & 503.0 &  \\
\quad \changedSecond{Deviance Difference} & 304.0 &  & 441.0 &  & 515.0 &  \\
	\hline
%\multicolumn{10}{c}{$^{***}p<0.001$, $^{**}p<0.01$, $^*p<0.05$}
\end{tabular}
\caption{Linear mixed effects models for reading time measures on the
  Dundee corpus, with model comparisons between base model and models
  including different types of surprisal.
	\changedSecond{For each coefficient, we show estimates and standard errors.}
	Boldfacing indicates model
  coefficients that are significant predictors. All model comparisons
  are significant at $p < 2.2 \cdot 10^{-16}$ using a $\chi^2$ test.}
\label{tab:neat-predicts-dundee}
\end{center}
\end{table}

%%%%%%%%%%%%%%%%%%%%%%%%%%%%%%%%%%%%%%%%%%%%%%%%%%%%%%%%%%%%%%%%%%%%%%%%%%%%%%%%
\section{Experiment~1}
\label{sec:exp1}

In Modeling Study~1 we successfully evaluated NEAT, our model of
reading based on the Tradeoff Hypothesis, against a corpus of human
eye-tracking data. We were able to show that human fixation sequences
are predicted by NEAT's measure of fixation probability, while human
reading time is predicted by it measure of restricted surprisal.

Recall that the version of NEAT used in Modeling Study~1 was based on
the assumption that the task that the model needs to solve is to build
up a representation of the text that allows it reconstruct the input
accurately. NEAT therefore learns to reconstruct an input sequence as
accurately as possible, while reading as economically as possible,
i.e., fixating as few words as it can.

We will now turn to an important prediction that can be derived from
the Tradeoff Hypothesis. If reading behavior is the consequence of a
tradeoff between task accuracy and reading economy, then this predicts
that the tradeoff will change when the reading task changes. More
specifically, if a reader is given a task that requires them to pay
particular attention to certain aspects of the text, then their
eye-movements will change in a way that is optimal for this
task. Experimental evidence for this comes, for instance, from studies
comparing proofreading and normal reading \citep{Schotter:ea:14}, see
Section~\ref{sec:tradeoff} for details.

In the following, we will present an eye-tracking experiment that
tests this prediction of the Tradeoff Hypothesis.  We investigate a
specific task, viz., reading a text in order to find the answer to a
question, and manipulate how much readers know about the task. In one
condition (No Preview), participants first read the text and then
answer a question about it; in the second condition (Preview), they
first see the question, then read the text, and then answer the
question. In the No Preview condition, readers have no idea what the
answer will look like, and their reading behavior should be similar to
normal reading: \changed{they try to build up a representation of the
  text that is as complete as possible,} as the question can be about
anything they have read. In the Preview condition, on the other hand,
readers know what type of information to seek, allowing them to read
faster and skip more words. In the Preview condition, we also expect
them to focus more on answer-relevant words (e.g.,~named entities),
and expect to see increased reading time on these words.

\subsection{Methods}
\label{sec:exp1:methods}

\subsubsection{Participants}

The experimental protocol was approved the ethics committee of the
School of Informatics at the University of Edinburgh.  Twenty-two
members of the University community took part in the experiment after
giving informed consent. They were paid \pounds{}10 for their
participation. All participants had normal or corrected-to-normal
vision and were self-reported native speakers of English.

\subsubsection{Materials}

Twenty newspaper texts were selected from the DeepMind question
answering corpus~\citep{hermann_teaching_2015}.  Ten texts were taken
from the CNN section of the corpus and the other ten texts from the
Daily Mail section.  Texts were selected so that they were comparable
in length and represented a balanced selection of topics.  Text
lengths ranged from 149 to 805 words (mean 323 words).  Two additional
texts were selected as practice items.  All texts were selected from
the test partition, which we did not use for training any of
the neural models described in this article.

For each text, a question and the correct answer were selected from
the corpus.  In the DeepMind corpus, questions are formulated as
sentences with a blank to be completed with a named entity so that a
statement implied by the text is obtained. An example is the following
question for the text in Figure~\ref{fig:heatmap-human}:

\ex. A random sample from a\ \ \_\_\_\_\_\_\_\_\_\_\ \ store tested
     positive for Listeria monocytogenes.
     
Questions were selected so that the correct answer does not occur at
the beginning of the text. The correct answer in this case is
\textit{Michigan}. All named entities are marked up in the DeepMind
corpus, a fact that we will use for our analysis later.

For each text, three incorrect answers (distractor) were created
(these are not included in the DeepMind corpus). The distractors were
also named entities, chosen so that correctly answering the question
would likely be impossible without reading the text. In the present
example, the distractors were names of other US states.

\subsubsection{Procedure}

The experiment included two conditions: Preview and No Preview. In the
Preview condition, participants first read the question, then they
read the text, and then they saw the question again with four answer
choices and had to select one answer. In the No Preview condition, the
question was not presented at the beginning of the trial, only after
the text had been read.

The design was between-groups, i.e., each participant took part either
in the Preview or No Preview version of the experiment. There were
10 participants in the Preview group and 10
participants in the No Preview group. In both groups, participants
first received written instructions (appropriate for their condition)
and went through two practice trials whose data was discarded. Then,
each participant read and responded to all 20 items (texts with
questions and answer choices); the items were the same for all
participants (modulo question preview), but were presented in a new
random order for each participant. The order of the answer options was
randomized for each condition, but the same answer order was used for
all participants in a given condition.

The experiment was conducted in a sound-attenuated room, where the
text was presented on a 24~inch LCD screen, in a Lucida Sans
Typewriter font with a fontsize of 20 points, line spacing of 20
points, and x- and y-offsets of 113 points.  Each trial consisted of
the question preview (in the Preview condition only; without answer
choices), presented on its own page. After a button press, the text
was displayed; texts were between one and five pages long (mean 2.1
pages), where each page contained up to eleven lines with about 80
characters per line. To get to the next page, and at the end of the
text, participants again had to press a button. After the last page of
text, the question was displayed, together with the four answer
choices, on a separate page. Participants had to press one of four
buttons to select an answer.

Eye-movements were recorded using an Eyelink 2000 tracker manufactured
by SR~Research (Ottawa, Canada). The tracker recorded the dominant eye
of the participant (as established by an eye-dominance test) with a
sampling rate of 2000~Hz. The participant was positioned about 60~cm
away from the screen, and a head rest was used to minimize head
movements. Before the experiment started, the tracker was calibrated
using a nine-point calibration procedure. At the start of each trial,
a fixation point was presented and drift correction was carried
out. Throughout the experiment, the experimenter monitored the
accuracy of the recording and carried out additional calibrations as
necessary. Button presses were collected using a USB game pad.

\subsubsection{Data Analysis}
\label{sec:data_analysis}

Drift in the vertical position of fixations was corrected
automatically. For this, we used custom-made software that adjusts the
vertical position of the nine calibration points, accordingly moving
recorded fixations. On each trial, the software minimizes a linear
combination of the squares of:
\begin{enumerate}
\item for each calibration point, the Euclidean distance from the
  recorded position;
\item the number of fixations not falling on any line of the text;
\item the number of pairs of successive fixations assigned to
  different lines of the text;
\item for each fixation falling within a line, the vertical distance
  from the center of that line;
\item for each fixation falling above the first line or below the last
  line, the vertical distance to the first or last line.
\end{enumerate}
We selected the coefficients for these five factors manually so as to
optimize the correction on a number of selected trials.  The software
only adjusted vertical positions; the horizontal positions of the
calibration points and thus of the fixations was left unchanged.

We also pooled short, contiguous fixations as follows: fixations of
less than 80~ms were incorporated into larger fixations within one
character, and any remaining fixations of less than 40~ms were
deleted. Readers do not extract much information during such short
fixations \citep{Rayner:Pollatsek:89}.

For data analysis, each word in the text was defined as a region of
interest. Punctuation was included in the region of the word it
followed or preceded without intervening whitespace. If a word was
preceded by a whitespace, then the space was included in the region
for that word.

We report data for the following eye-movement measures in the critical
and spill-over regions. \emph{First fixation duration} is the duration
of the first fixation in a region, provided that there was no earlier
fixation on material beyond the region.  \emph{First pass time} (often
called gaze duration for single-word regions) consists of the sum of
fixation durations beginning with this first fixation in the region
until the first saccade out of the region, either to the left or to
the right.  \emph{Total time} consists of the sum of the durations of
all fixation in the region, regardless of when these fixations occur.
\emph{Fixation rate} measures the proportion of \changed{subjects who
  fixated (rather than skipped) the region} on first-pass reading.

For first fixation duration and first pass time, no trials in which
the region is skipped on first-pass reading (i.e.,~when first fixation
duration is zero) were included in the analysis. For total time, only
trials with a non-zero total time were included in the analysis.

\subsection{Results}
\label{sec:exp1:results}

\begin{table}[]
\begin{center}
\begin{tabular}{ldddddddddd}
\hline
& \mc{1}{c}{\multirow{2}{*}{No Preview}} &  \mc{1}{c}{\multirow{2}{*}{Preview}} & \mc{1}{c}{No Preview,} & \mc{1}{c}{No Preview,} & \mc{1}{c}{Preview,} & \mc{1}{c}{Preview,}  \\ 
& & & \mc{1}{c}{before answer} & \mc{1}{c}{after answer}  & \mc{1}{c}{before answer} & \mc{1}{c}{after answer}  \\
\hline
First fixation &214.1 & 194.1 & 215 & 213.6 & 193.2 & 194.7 \\ 
First pass &257.3 & 213.2 & 259.6 & 256 & 213.2 & 213.3 \\ 
Total time &317.6 & 262.5 & 329.8 & 310.3 & 266.5 & 259.9 \\ 
Fixation rate &0.52 & 0.32 & 0.51 & 0.52 & 0.32 & 0.32 \\ 
%First fixation &214.1 & 194.1 & 193.2 & 194.7 \\ 
%First pass &257.3 & 213.2 & 213.2 & 213.3 \\ 
%Total time &317.6 & 262.5 & 266.5 & 259.9 \\ 
%Fixation rate &0.52 & 0.32 & 0.32 & 0.32 \\ 
\hline
Accuracy       & \mc{1}{c}{70\%} & \mc{1}{c}{89\%} \\
\hline
%rp (w/o NA) & 463.9 & 362.5 & 359.5 & 363.5 \\
\end{tabular}
\caption{Left: mean fixation rate, reading time, and question
  accuracies by conditions in our reading experiment. Right: the same
measures separately for words that
  occur before and after the correct answer in the
  text.}\label{tab:stat}
\end{center}
\end{table}

\subsubsection{Tradeoff between Accuracy and Economy}
\label{sec:tradeoff:res}

Table~\ref{tab:stat} shows descriptive statistics for the reading time
measures first fixation duration, first pass time, and total time. We
also report fixation rate, i.e., the proportion of words that were
fixated rather than skipped.

While the fixation rate was 0.52 in the No Preview condition, it
dropped to 0.32 in the preview condition.  Similarly, all reading time
measures were substantially lower in the Preview condition: we observe
a 20~ms reduction in first fixation duration, and reductions of 44~ms
and 55~ms in first pass duration and total time, respectively. (We
statistically analyze these differences using mixed effects models
below.) Note also that fixation rate was substantially lower than in
the Dundee corpus.

It is possible this result is simply due to the participants in the
Preview condition being strategic: once they have found the answer in
the text, they only read the rest of the text superficially, or even
skip it completely. This would result in reduced overall reading time
and fixation rate when averaging across the text as a whole. The
right half of Table~\ref{tab:stat} presents reading time and fixation
rate for the Preview condition separately for the words in the text
that occur before and after the answer. The measures in both cases are
indistinguishable (except for a small reduction in total time). This
is evidence that participants read all of the text in the same way,
and do not adopt a special strategy once they have found the answer.

Turning now to question answering accuracy, we found an accuracy of
70\% in the No Preview condition, rising to 89\% in the Preview
condition.  This difference was significant ($\beta = 2.05$,
$\mathit{SE} = 0.69$, $p = 0.0027$) in a logistic mixed effects model
with text and participant as random effects and condition as fixed
effect, and the appropriate random intercepts and slopes
\changed{(i.e., a by-text slope for condition)}.  The question
accuracies show that participants were reading the texts attentively,
performing substantially above chance (25\% accuracy) in both
conditions, despite the low fixations rate observed in this
experiment.

\begin{figure}[]
	\centering
	No Preview
\vspace{2ex}

\includegraphics[width=.9\textwidth]{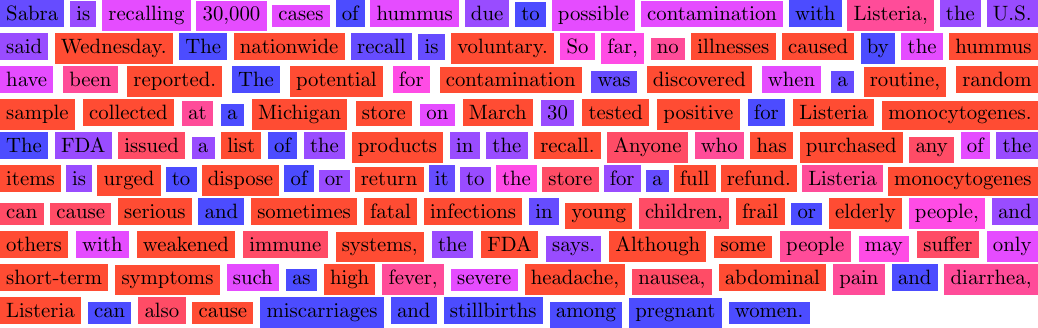}

\vspace{2ex}
Preview
\vspace{2ex}

\includegraphics[width=.9\textwidth]{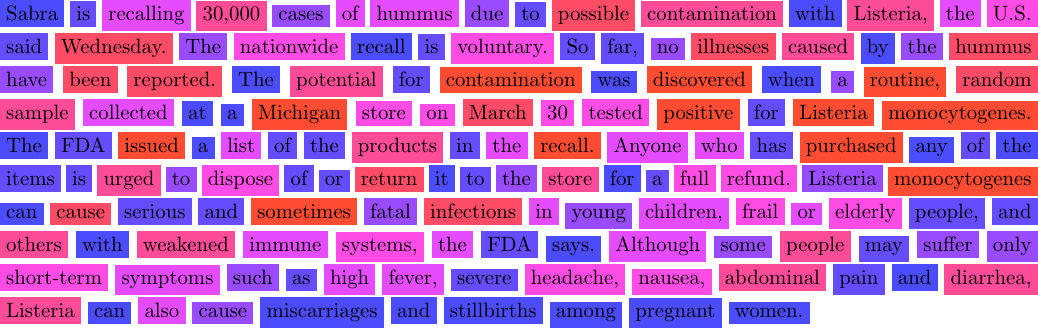}

\vspace{2ex}

\caption{
	Heatmap visualizing human fixation probabilities in
  Experiment~1, averaged over all participants per condition. Top: No
  Preview, bottom: Preview.  The color gradient denotes the fixation
  probability for a word, ranging from blue (0.0) to red (1.0). The
  question was \textit{A random sample from
    a\ \ \_\_\_\_\_\_\_\_\_\_\ \ store tested positive for Listeria
    monocytogenes}. The correct answer was \textit{Michigan}, which
  occurs in the fourth line of the text.}\label{fig:heatmap-human}
\end{figure}

A sample visualization of skipping behavior in the two conditions is
shown in Figure~\ref{fig:heatmap-human}.  Overall fixation rate is
higher in the No Preview condition (top) than in the Preview condition
(bottom).  In the No Preview condition, most content words were
fixated at least by some participants, and longer words were fixated
by most participants. This contrasts with the Preview condition, where
only a few words have high fixation rate, including the correct
answer (\textit{Michigan}, in the fourth line of the text).

Based on this observation, we wanted to test whether spending more
time reading the answers in the text enables participants to answer
the questions more accurately. We therefore built a logistic mixed
effects model predicting whether a question was answered correctly
based on (1)~condition (Preview or No Preview), and (2)~total time
spent on all the occurrences of the correct answer in the text,
centered and scaled to unit variance.  The model included text and
participant as random effects, and the appropriate random slopes
\changed{(i.e., by-text slope for condition, and by-text and
  by-participant slopes for total time)}.  We conducted a Bayesian
logistic analysis using \texttt{brms} \citep{buerkner2017brms}, as
frequentist models were numerically unstable.  Reading time was a
significant predictor ($\beta = 1.22$, $\mathit{SE} = 0.3434$,
$95\%\ CrI = [0.39, 2.22]$), with higher reading time increasing the
probability of answering correctly.  The main effect of condition was
also significant ($\beta = 2.00$, $\mathit{SE}=0.74$, $95\%\ CrI =
[0.70,3.67]$).  There was an interaction, such that the effect of
reading time was stronger in the Preview condition ($\beta = 1.81$,
$\mathit{SE}=0.98$, $95\%\ CrI = [0.04, 3.90]$).

\subsubsection{Mixed Effects Analyses}

\begin{table}[]
\begin{center}
%\begin{tabular}{l@{}d@{}d@{~~}l@{~~}d@{}d@{~~}l@{~~}d@{}d@{~~}l@{~~}d@{}d@{~~~~}l@{}}
\begin{tabular}{l@{}r@{}r@{~~}r@{}r@{~~}r@{}r@{~~}r@{}r@{~~~~}}
\hline
& \mc{2}{c}{First Fixation}& \mc{2}{c}{First Pass} & \mc{2}{c}{Total
  Time} & \mc{2}{c}{Fixation Rate} \\ \hline
(Intercept)                   &\textbf{201.60}&(4.71)  &\textbf{227.17}&(6.84)  &\textbf{276.30}&(7.61)  &\textbf{-0.35}&(0.16)   \\
Task                     &\textbf{20.09}&(9.40)  &\textbf{48.53}&(13.62) &\textbf{67.52}&(15.10) &\textbf{0.98}&(0.31)   \\
IsCorrectAnswer               &-2.31   &(3.79)  &\textbf{-27.84}&(10.49) &8.21    &(16.81) &0.21    &(0.13)   \\
IsNamedEntity                 &-2.73   &(1.42)  &\textbf{-10.14}&(2.86)  &4.96    &(4.38)  &0.05    &(0.06)   \\
WordLength                    &0.40    &(0.24)  &\textbf{8.11}&(0.38)  &\textbf{13.87}&(0.58)  &\textbf{0.29}&(0.01)   \\
LogWordFreq                   &\textbf{-6.30}&(0.52)  &\textbf{-7.28}&(0.28)  &\textbf{-40.84}&(1.26)  &\textbf{-0.65}&(0.04)   \\
PositionText                  &0.74    &(0.45)  &-1.10   &(0.76)  &\textbf{-9.87}&(1.15)  &0.02    &(0.1)    \\
Surprisal                     &\textbf{1.27}&(0.19)  &\textbf{1.43}&(0.30)  &\textbf{3.39}&(0.45)  &0.01    &(0.01)   \\
\hline
IsCorrectAnswer:Surprisal     &        &        &{\textbf{-5.20}}&(1.98)  &\textcolor{black}{\textbf{-16.07}}&(3.03)  &        &         \\
IsCorrectAnswer:WordLength    &        &        &        &        &\textcolor{black}{\textbf{18.58}}&(4.56)  &        &         \\
IsNamedEntity:Surprisal       &        &        &{\textbf{-1.96}}&(0.68)  &\textcolor{black}{\textbf{-3.62}}&(1.01)  &{\textbf{0.02}}&(0.01)   \\
IsNamedEntity:WordLength      &        &        &        &        &\textcolor{black}{\textbf{6.27}}&(1.53)  &        &         \\
LogWordFreq:IsCorrectAnswer   &        &        &{\textbf{-12.69}}&(2.48)  &\textcolor{black}{\textbf{-96.79}}&(12.30) &        &         \\
LogWordFreq:IsNamedEntity     &        &        &{\textbf{-6.53}}&(0.76)  &\textcolor{black}{\textbf{-44.01}}&(3.61)  &{\textbf{0.21}}&(0.05)   \\
LogWordFreq:Surprisal         &\textbf{0.52}&(0.21)  &        &        &        &        &        &         \\
LogWordFreq:WordLength        &\textbf{1.06}&(0.23)  &        &        &        &        &\textbf{0.05}&(0.01)   \\
LogWordFreq:PositionText      &        &        &\textbf{0.81}&(0.25)  &\textbf{6.11}&(1.24)  &        &         \\
PositionText:IsCorrectAnswer  &        &        &{\textbf{-31.76}}&(8.89)  &\textcolor{black}{\textbf{-115.03}}&(14.20) &        &         \\
PositionText:IsNamedEntity    &        &        &        &        &\textcolor{black}{\textbf{-18.97}}&(3.52)  &        &         \\
Surprisal:WordLength          &\textbf{0.17}&(0.08)  &\textbf{0.34}&(0.13)  &        &        &        &         \\
\hline
Task:IsCorrectAnswer     &\textcolor{black}{\textbf{-12.97}}&(5.84)  &        &        &\textcolor{black}{\textbf{-186.23}}&(12.06) &\textcolor{black}{\textbf{-0.31}}&(0.16)   \\
Task:IsNamedEntity       &        &        &        &        &        &        &\textcolor{black}{\textbf{-0.19}}&(0.09)   \\
Task:LogWordFreq         &        &        &\textcolor{black}{\textbf{-4.28}}&(0.39)  &\textcolor{black}{\textbf{-18.46}}&(1.52)  &\textcolor{black}{\textbf{-0.14}}&(0.07)   \\
Task:Surprisal           &        &        &        &        &        &        &-0.02   &(0.01)   \\
Task:WordLength          &\textcolor{black}{\textbf{-0.89}}&(0.35)  &\textcolor{black}{\textbf{4.11}}&(0.55)  &\textcolor{black}{\textbf{5.11}}&(0.73)  &0.04    &(0.03)   \\
Task:PositionText        &        &        &        &        &        &        &0.06    &(0.08)   \\
\hline
%\multicolumn{13}{c}{$^{***}p<0.001$, $^{**}p<0.01$, $^*p<0.05$}
\end{tabular}
\vspace{1ex}
\caption{Mixed effects models for the data from Experiment~1.
  Task is coded as $-0.5$ (Preview) vs. $+0.5$ (No Preview).
	\changedSecond{Binary interactions were identified automatically using forward model selection, as described in the text.}
	For
  first fixation, first pass and total time, we report the estimated
	coefficient and \changedSecond{its standard error} as obtained using linear
  mixed effects models. For fixation rate, the dependent variable is
  binary (word fixated or not), so we fitted Bayesian logistic mixed
	effects models and report the estimated coefficient and its \changedSecond{posterior standard}
  deviation.  Effects were interpreted as significant when $|t|>2$
  (reading time) or the posterior probability that the coefficient
  has the opposite sign was $<0.05$ (fixation rate).  Continuous
  predictors were centered and scaled to have unit standard
  deviation.}
\label{tab:reading-times-lme:exp1}
\end{center}
\end{table}

In order to more comprehensively analyze the effect of our
experimental manipulation on eye-tracking measures, we fitted a series
of mixed effects models to the data. We include not only task
(Preview or No Preview, coded as $-0.5$ and $+0.5$, respectively) as a
fixed factor in our models, but also the following word-based factors,
which allow us to analyze in more detail how the question-answering
task influences reading strategy:
\begin{enumerate}
\item LogWordFreq: log-transformed word frequency, computed from the
	\changed{training set of the DeepMind corpus (230 million words} of
  newstext);
\item WordLength: length of the word in characters, residualized with
	respect to log word frequency \changed{(see SI Appendix, Section~3)};
\item PositionText: the position of the word in the text, counted from
  the first word of the text;
\item Surprisal: $-\log P(w_n|w_{1 \dots n-1})$ computed using a
  recurrent neural network language model trained on the training
  set of the DeepMind corpus. This was residualized with respect to
		log word frequency \changed{(see SI Appendix, Section~3).  After residualizing}, we set surprisal to zero
  for those words that are outside of the vocabulary of the neural
		language model (2.8\% of tokens), as it does not compute meaningful surprisal
  estimates for those.
\item IsNamedEntity: whether the word is part of a named entity (i.e.,
  a potential answer), coded as $-0.5$ for no and $+0.5$ for yes.
\item IsCorrectAnswer: whether the word is part of the correct answer
  to the question for this text, again coded as $-0.5$ for no and
  $+0.5$ for yes, and then residualized with respect to IsNamed\-Entity
  (only named entities can belong to the correct answer). % and the
\end{enumerate}
For first fixation, first pass, and total time, we fitted mixed
effects models using the R~package
\texttt{lme4}~\citep{bates-fitting-2015-1}. These models included
random intercepts for participants and items and their outputs are
shown in Table~\ref{tab:reading-times-lme:exp1}. Models with random
slopes did not converge when using \texttt{lme4}. We therefore also
built Bayesian linear mixed effects models with all appropriate random
slopes  using \texttt{brms} \citep{buerkner2017brms}, both for reading
times and for log-transformed reading time; the results of these
models broadly agree with the non-Bayesian models, and are given in
SI Appendix, Section~2.

For fixation rate, the dependent variable was binary (word fixated or
skipped). We found that fitting logistic mixed models with
\texttt{lme4} gave unstable results, so we fitted Bayesian logistic
mixed effects models with participants and items as random effects and
all appropriate random slopes using \texttt{brms}.  The fitted model
for fixation rate is also shown in
Table~\ref{tab:reading-times-lme:exp1}.
See SI Appendix, Section~2 for details.

For first fixation, first pass, and total time, only fixated words
were included in the analysis. All predictors were centered.
Continuous predictors were scaled to unit standard deviation, and
categorical predictors had a difference of one between their two
levels.  The coefficient of a predictor in these models can therefore
be interpreted as the change in milliseconds per standard deviation of
the predictor (continuous), or the change (in milliseconds) induced by
going from one level to the other (categorical predictors). 
\changedSecond{In order to select which binary interactions to include in the models,} we
conducted forward model selection with a $\chi^2$ test as described in
Section~\ref{sec:study1:results:rt}.
We removed the first token of each text from analysis.

\subsubsection{Task-independent Effects}

We will first discuss the main effect found in our mixed effects
analysis, see the first section of
Table~\ref{tab:reading-times-lme:exp1}. The results show main effects
\changedSecond{of log word frequency (all measures) and word length (first pass, total time, fixation rate)}, indicating that
infrequent words and long words are read more slowly and are more
likely to be fixated, independent of task.
\changedSecond{There is also a significant effect of named entity status
on first pass. Words that are part of
a named entity (i.e.,~a potential answer) are read faster in this measure.}
\changedSecond{We also observe a main effect of whether a word is part of the correct
answer (decreased first pass time) and of text position:
words later in the text have lower total time.}
There is also a significant main effect of surprisal on all reading
time measures: more surprising words are read more slowly. This is
consistent with previous work on surprisal in reading time corpora
\citep{demberg_data_2008}. There is no effect of surprisal on fixation
rate.

We now turn to interactions that do not involve task, which are
given in the second section of Table~\ref{tab:reading-times-lme:exp1}.
We find an interaction of word length and word frequency in first
fixation and fixation rate, which is a standard reading
time effect.  We also observe that the position of a word in the text
interacts with its frequency (in first pass and total time), which is again a
standard reading time effect.
\changedSecond{There are negative interactions of text position with named entity
status (total time) and answer status (first pass and total time), which indicates that named entities are read
faster later in the text, presumably because a named entity is less
likely to be the correct answer later in the text, as the answer has
already been encountered by then.}

We also find four interactions involving surprisal. There is a positive
interaction of surprisal and word length in first fixation and first pass, indicating that surprisal has a bigger effect on
longer words than on shorter words.
\changedSecond{An interaction between log word frequency and surprisal is found in first fixation.}
\changedSecond{There are also interactions of surprisal and named entity
status, and between surprisal and answer status. Words that belong to named entities are less affected by surprisal than other words in the reading time measures, but more than other words in fixation rate.}
\changedSecond{Relatedly, we find interactions between answer status and log word frequency (first pass and total time), and between answer status and word length (total time), showing that frequency and length effects are stronger on the correct answer than on other words.}

We also find interactions of named entity status with log word
frequency (first pass, total time, and fixation rate).  This indicates
that the effect of log word frequency is more pronounced on named
entities in reading time, and less pronounced in fixation rate.

\subsubsection{Task-dependent Effects}

%%%%%%%%%%%%%%%%%%%%%%%%%%%%%%%%%%%%%%%%%%%%%%%%%%%%%%%%%%%%%%
% The code collecting the figures is auto-generated from the results table
%%%%%%%%%%%%%%%%%%%%%%%%%%%%%%%%%%%%%%%%%%%%%%%%%%%%%%%%%%%%%%
%\input{../study2/analysis/analysis/mixed_models/print_models/output/putTogetherModels_printFiguresCode.py.tex}

\begin{figure}[]
\begin{center}

\includegraphics[width=.5\textwidth]{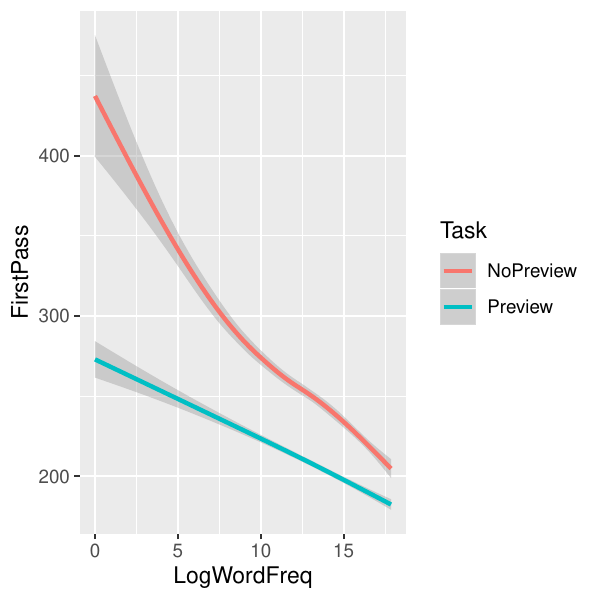}%
\includegraphics[width=.5\textwidth]{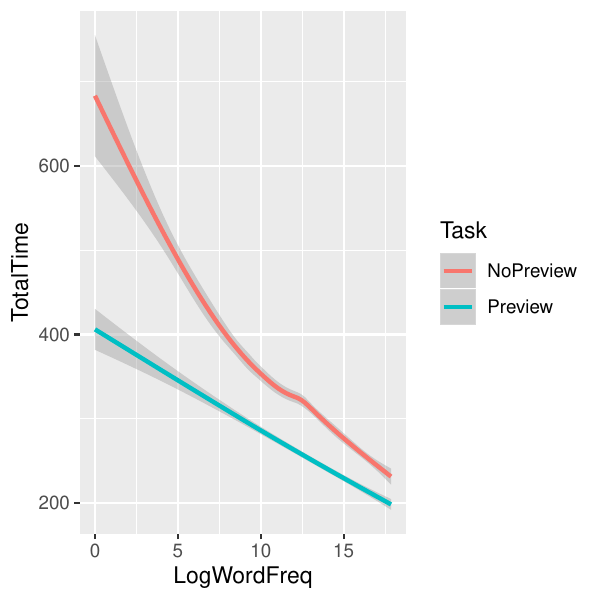}%

\includegraphics[width=.5\textwidth]{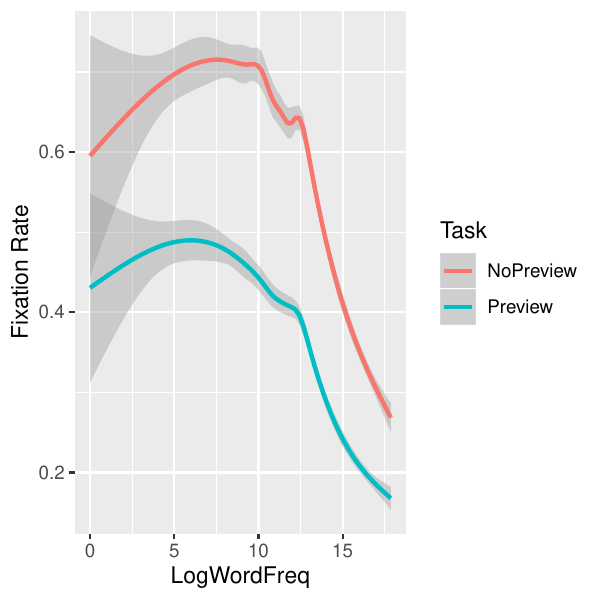}%
\end{center}
        \caption{Interaction between task and log word frequency in
          first pass duration, total time and fixation rate.
 We show GAM-smoothed means with 95\% confidence intervals.
}
\label{fig:cond:logwf}
\end{figure}

\begin{figure}[]
\begin{center}

\includegraphics[width=.5\textwidth]{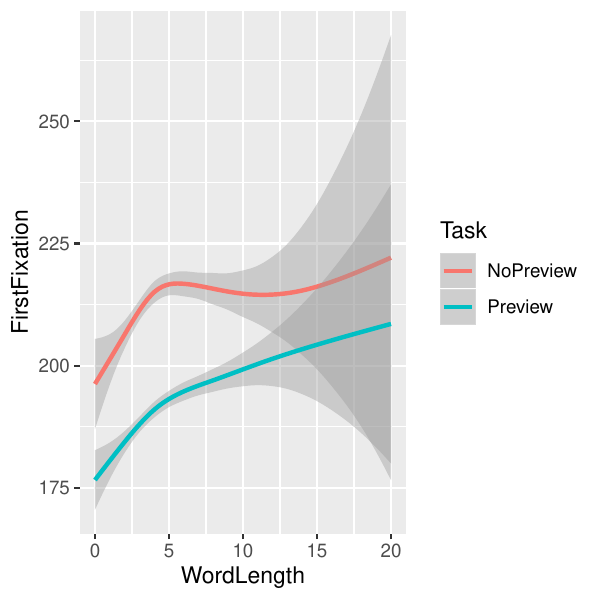}%
\includegraphics[width=.5\textwidth]{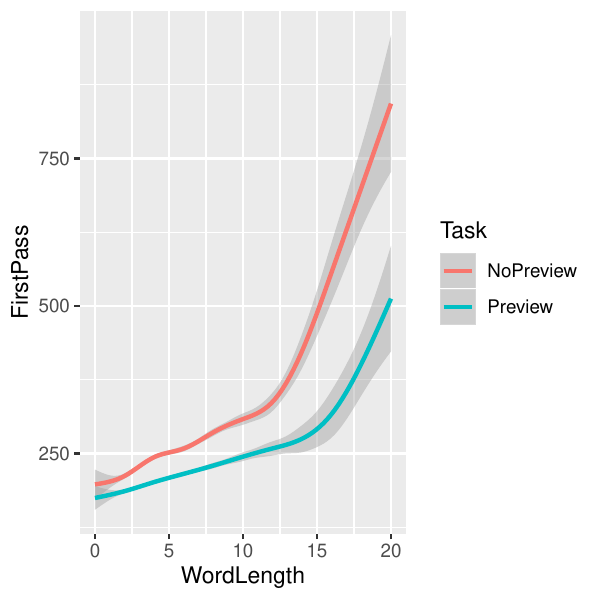}%

\includegraphics[width=.5\textwidth]{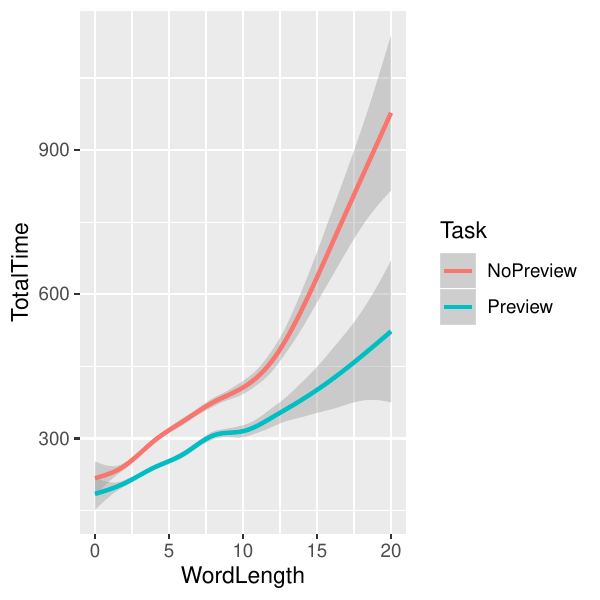}%
\end{center}
        \caption{Interaction between task and word length in first fixation duration, first pass duration and total time.
 We show GAM-smoothed means with 95\% confidence intervals.
}
\label{fig:cond:wlen}
\end{figure}

\begin{figure}[]
\begin{center}

\includegraphics[width=.5\textwidth]{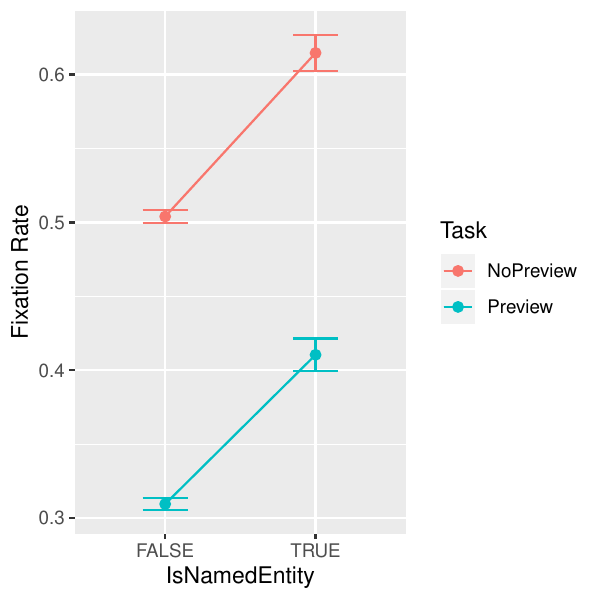}%
\end{center}
        \caption{Interaction between task and named entity status in
          fixation rate.
}
\label{fig:cond:ne}
\end{figure}

\begin{figure}[]
\begin{center}

\includegraphics[width=.5\textwidth]{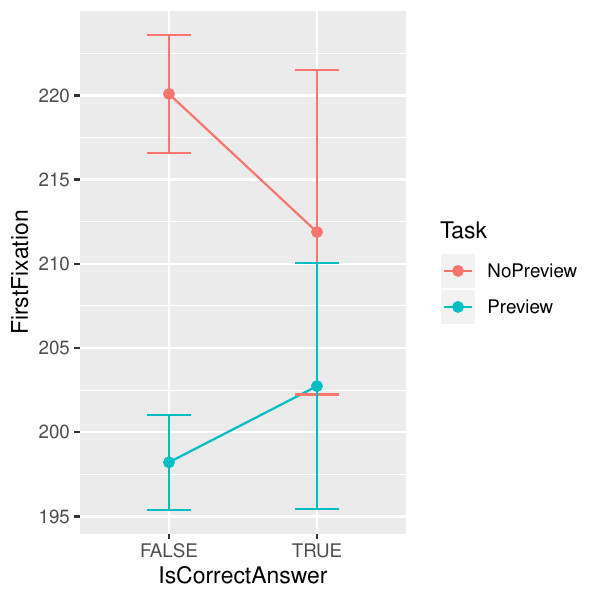}%
\includegraphics[width=.5\textwidth]{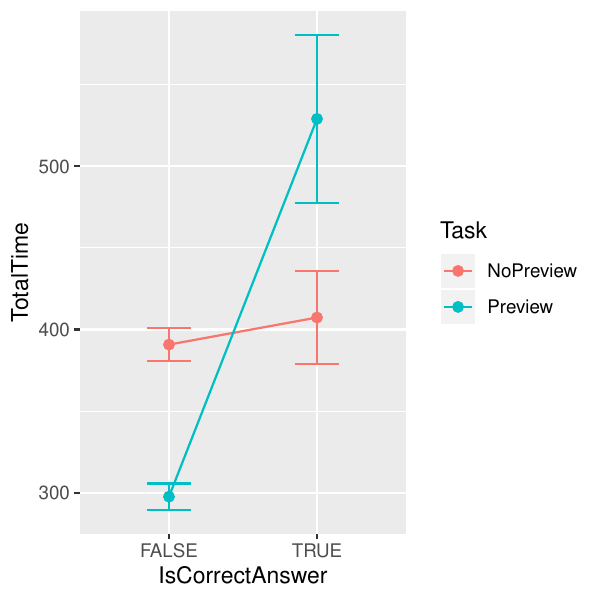}%

\includegraphics[width=.5\textwidth]{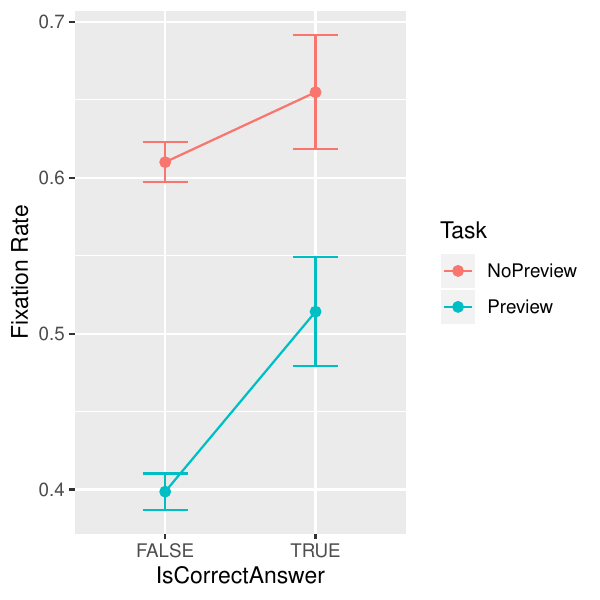}%
\end{center}
        \caption{Interaction between task and answer status in first
          fixation duration, total time and fixation rate.
We only show datapoints belonging to named entities, as answer status is only relevant for those. 
}
\label{fig:cond:corrans}
\end{figure}

\changedSecond{As predicted by the Tradeoff Hypothesis,} we find a
significant main effect of task in all measures: first fixation, first pass, total
time, and fixation rate are all higher in the No Preview condition,
confirming the observations we made based on the descriptive
statistics in Table~\ref{tab:stat}.

We now turn to the third section of
Table~\ref{tab:reading-times-lme:exp1}, which shows interactions
involving task. \changedSecond{As explained in Section~\ref{sec:exp1},
  we expect readers in the Preview condition to focus more on
  answer-relevant words (e.g.,~named entities), which means we should
  see increased reading time on these words.}  We visualize the
significant interactions in
Figures~\ref{fig:cond:logwf}--\ref{fig:cond:corrans}.

The results show a significant negative interaction between task
and word frequency in first pass, total time, and fixation rate (see
Figure~\ref{fig:cond:logwf}). This means that the word frequency
effect is less pronounced in the Preview condition. We observe a
similar effect for word length (see Figure~\ref{fig:cond:wlen}): the
positive interaction of task and word length in first pass and
total time indicates that word length has a weaker effect on reading
time in the Preview condition; though there is a negative interaction
in first fixation.  We hypothesize that when readers have preview of
the question, they adopt a strategy in which the allocation of
attention (i.e.,~the reading time) depends more heavily on information
extracted from the question, and is less reliant on low-level factors
such as word frequency and word length.

We also find a \changedSecond{negative} interaction of task and named entity
status in fixation rate (see Figure~\ref{fig:cond:ne}). 
\changedSecond{This indicates that the overall task effect on fixation rate is reduced for named entities, potentially because readers focus more on such highly answer-relevant words in the Preview condition.}

Finally, we observe an interaction between task and
IsCorrectAnswer in first fixation, total time, and fixation rate (see
Figure~\ref{fig:cond:corrans}).  This is a large effect: words
occurring in the correct answer are read for an extra \changedSecond{186~ms} in the
Preview condition. Again, this provides clear evidence of a task
effect.  Readers modify their reading strategy when they have access
to information about the question: their main goal is to find the
answer in the text, and if they have found words that could belong to
the answer, they want to make sure of this, and spend a extra time
reading these words.

\subsection{Discussion}
\label{sec:exp1:discussion}

We found clear differences in reading time, fixation rate, and
accuracy between the two experimental conditions. In the Preview
condition, participants read faster and skipped more, but achieved a
higher accuracy, compared to the No Preview condition.  This is
consistent with our hypothesis that task has an effect on reading
behavior: the presence of the question at the start of a trial changes
the task from a normal reading task to an information-seeking
task. This in turn changes the economy--accuracy tradeoff (as
predicted by NEAT): even with reduced reading time and increased
skipping rate (more economical reading), readers achieve increased
accuracy.

In addition to the main effect of task, we also found main effects of
word frequency, word length, text position, and surprisal, which are
well known from the reading time literature
\citep[e.g.,][]{demberg_data_2008}.
\changedSecond{Furthermore, we found words that are parts of named
  entities, particularly those that are part of the correct answer,
  are read faster in first pass reading.}

Finally, we uncovered a set of interactions between task (Preview or
No Preview) and word frequency, word length, named entity status, and
correct answer status.  \changedSecond{Across reading measures,
  effects of frequency and length were reduced in the Preview
  condition. As participants already know what kind of information to
  look for, their reading is affected more by the question, and less
  by low-level properties such a frequency.}
\changedSecond{In the Preview condition,} they also spend more time
reading words that occur in the answer, \changedSecond{potentially in
  order} to make sure that they have indeed found the right
answer. Again, this provides evidence for task effects, i.e., a change
in reading strategy depending on whether participants perform a task
that is similar to standard reading (No Preview), or an information
seeking task (Preview).

Overall, this experiment demonstrated that reading behavior is finely
attuned to the reading task at hand. For instance, in the particular
task we evaluated, named entities play a crucial role, and we found
reading behavior that is adapted to this fact. The challenge now is to
devise a computational model that is able to capture both the main
effects and the complex pattern of interactions we observed in the
human reading data collected in this experiment.

%%%%%%%%%%%%%%%%%%%%%%%%%%%%%%%%%%%%%%%%%%%%%%%%%%%%%%%%%%%%%%%%%%%%%%%%%%%%%%%%
\section{Modeling Task Effects in NEAT}
\label{sec:neat_task}

\def\leftend{-7.7}
\def\interval{2.5}
\def\surpfactor{0.3}

\def\la{\leftend}
\def\lb{\leftend+\interval}
\def\lc{\leftend+2*\interval}
\def\ld{\leftend+3*\interval}
\def\le{\leftend+4*\interval}

\def\laa{\leftend+0.1*\interval}
\def\lba{\leftend+1.1*\interval}
\def\lca{\leftend+2.1*\interval}

\def\surpcolor{black}
\def\fixcolor{black}

\begin{figure}[]
\centering
\begin{tikzpicture}[%
  % common options for blocks:
  block/.style = {draw, fill=blue!30, align=center, anchor=west,
              minimum height=0.65cm, inner sep=0},
  % common options for the circles:
  ball/.style = {circle, draw, align=center, anchor=north, inner sep=0}]

\node[rectangle,draw,text width=1.5cm,anchor=base] (Q) at (\leftend+4+0.3*\interval,1) {$q_1\ \ q_2\ \ q_3$};

\node[rectangle,draw=none,text width=0.3cm,anchor=base] (W1) at (\leftend+0.3*\interval,-1) {$w_1$};
\node[rectangle,draw=none,text width=0.3cm,anchor=base] (W2) at (\leftend+1.3*\interval,-1) {$w_2$};
\node[rectangle,draw=none,text width=0.3cm,anchor=base] (W3) at (\leftend+2.3*\interval,-1) {$w_3$};

{\node[rectangle,draw,text width=0.3cm,anchor=base,\fixcolor] (A1) at (\leftend+\interval,-1) {A};}
{\node[rectangle,draw,text width=0.3cm,anchor=base,\fixcolor] (A2) at (\leftend+2*\interval,-1) {A};}
{\node[rectangle,draw,text width=0.3cm,anchor=base,\fixcolor] (A3) at (\leftend+3*\interval,-1) {A};}

{\node[rectangle,draw,text width=0.3cm,anchor=base] (R0) at (\leftend,-3) {$R_0$};}
{\node[rectangle,draw,text width=0.3cm,anchor=base] (R1) at (\leftend+\interval,-3) {$R_1$};}
{\node[rectangle,draw,text width=0.3cm,anchor=base] (R2) at (\leftend+2*\interval,-3) {$R_2$};}
{\node[rectangle,draw,text width=0.3cm,anchor=base] (R3) at (\leftend+3*\interval,-3) {$R_3$};}

{\node[rectangle,draw=none,text width=0.3cm,anchor=base] (RW1) at (\leftend+\interval,-2) {$w_1$};}
	{\node[rectangle,draw=none,text width=0.9cm,anchor=base] (RW2) at (\leftend+2*\interval,-2) {\text{\footnotesize{\textsc{skipped}}}};}
{\node[rectangle,draw=none,text width=0.3cm,anchor=base] (RW3) at (\leftend+3*\interval,-2) {$w_3$};}

%\node[rectangle,draw,text width=8cm,anchor=base] (M) at (\leftend+3.8,-4) {Answer Selection};

\draw (\leftend+2,-4) rectangle (\leftend+8,-4.7) node[pos=.5,inner sep=4mm,outer sep=0 ] (M) {Answer Selection};

\draw[->] (R1.south) -- (M);
\draw[->] (R2.south) -- (M);
\draw[->] (R3.south) -- (M);
\draw[->] (Q.east) to [out=0,in=0] (\leftend+8,-4.35);

{\draw[->] (Q.south) to [out=250,in=70] (A1.north);}
{\draw[->] (Q.south) to [out=280,in=90](A2.north);}
{\draw[->] (Q.south) to [out=320,in=110](A3.north);}

{\draw[->] (R0.east) to [out=0,in=180] (R1.west);}
{\draw[->] (R1.east) to [out=0,in=180] (R2.west);}
{\draw[->] (R2.east) to [out=0,in=180] (R3.west);}

%{\draw[->] (R0.east) to [out=10,in=250]   (A1.225);}
%{\draw[->] (R1.east) to [out=10,in=250]   (A2.225);}
%{\draw[->] (R2.east) to [out=10,in=250]   (A3.225);}
%
{\draw[->] (W1.10) to [out=0,in=180] (A1.west);}
{\draw[->] (W2.10) to [out=0,in=180] (A2.west);}
{\draw[->] (W3.10) to [out=0,in=180] (A3.west);}

{\draw[->] (RW1.south) to [out=270,in=90] (R1.north);}
{\draw[->] (RW2.south) to [out=270,in=90] (R2.north);}
{\draw[->] (RW3.south) to [out=270,in=90] (R3.north);}

%\draw[->] (RW1.south) to [out=300,in=120] (M);
%{\draw[->] (RW2.south) to [out=,in=90] (M);}
%\draw[->] (RW3.south) to [out=240,in=60] (M);

{\draw[->] (A1.south) to [out=270,in=90] (RW1.north);}
{\draw[->] (A2.south) to [out=270,in=90] (RW2.north);}
{\draw[->] (A3.south) to [out=270,in=90] (RW3.north);}

\end{tikzpicture}

\hspace{1ex}

\caption{Architecture of the NEAT question answering model as applied to a text $w_1 w_2 w_3$ together with a question $q_1 q_2 q_3$.}
\label{fig:architecture:qa}
\end{figure}

In the version introduced in Section~\ref{sec:neat}, the NEAT model
assumes the following default task: during normal reading,
\changed{the reader tries to build up a representation of the text
  that enables them to later recall it as accurately as possible.}
The cognitive tradeoff that readers face is therefore one between
skipping as many words as possible (economy) and reconstructing the
input as well as possible (accuracy). In NEAT, this is implemented by
combining a reader module (which predicts the next word) with an
attention module (which decides whether to fixated the next word or
not), and a task module (which tries to reconstruct the input based on
the text representation computed by the reader).

The Tradeoff Hypothesis predicts that the model's behavior should
change if the reading task changes and necessitates a different
tradeoff between task accuracy and reading economy. The results from
the eye-tracking experiment reported in Section~\ref{sec:exp1} confirm
this prediction. We gave readers a question answering task and
manipulated whether they had access to the question before they read
the text or not (Preview or No Preview). The results clearly show that
preview has an effect on reading strategy, affecting both reading
times and skipping behavior. The NEAT model should be able to capture
this finding if we assume a different task module, viz., one that
performs question answering rather than input reconstruction. In this
section, we introduce a modified version of NEAT that has this
property and evaluate it against the eye-tracking data from
Experiment~1.

\subsection{Architecture}
\label{sec:neat_task:architecture}

The architecture of our revised version of NEAT is diagrammed in
Figure~\ref{fig:architecture:qa}.  For illustrative purposes, we
assume the input text consists only of the words $w_1, w_2, w_3$. As
further input, the model now receives a question, consisting of the
words $q_1, q_2, q_3$ in our example.  Like in the previous version of
NEAT, the reader module is a recurrent neural network which reads the
text in linear order and creates a memory representation, recording a
vector of neural activations at each word. These vectors are denoted
as $R_0, R_1, R_2, R_3$ in Figure~\ref{fig:architecture:qa}.

As before, the attention module $A$ decides for each input word $w_i$
whether to read or skip that word.  If $w_i$ it is read, then it is
shown to the reader and incorporated into its memory representation.
If it is skipped, then the reader is only shown a placeholder symbol
indicating that a word was skipped.

In the Preview condition, the attention module $A$ has access to the
question $q_1, q_2, q_3$ when making decisions about fixations and
skips.  In Figure~\ref{fig:architecture:qa}, this is marked by
connections going from the question to the attention module.  In the
No Preview condition, these connections are absent, and the attention
module $A$ has no access to the question while it is reading the input
text.

%After the text has been read, 
\changedSecond{The representations created by the
reader} are passed to the task module, which is a neural network which
matches the memory representation with the question and attempts to
select the correct answer. At this point, the question is available to
the model, independently of whether it is in the Preview or No
Preview condition.

The reader module and the task module can be trained using supervised
learning on the basis of a corpus of texts with questions and answers,
with the objective of maximizing question answering accuracy. As
before, the attention module is trained using reinforcement learning,
but now a question answering task rather than a reconstruction task
provides the reinforcement signal: during training, the module
generates fixation decisions, and passes the fixated words on to the
reader.  The parameters of the attention module
are then updated to upweight decisions that led to correct answers,
and downweight decisions that did not.

The reader module remains unchanged from our original version of NEAT
as introduced in Section~\ref{sec:neat} (but the hidden layer is
smaller, containing 128 memory cells only). However, changes are
necessary to the attention and task modules, which we will describe in
the following sections.

\subsubsection{Attention Module}
\label{sec:attention-module}

Our new version of NEAT is based on the \emph{NoContext} version of
the original NEAT model.  As in Modeling Study~1, we compute fixation
probabilities~$a_i$ of a word $w_i$ by applying a linear
transformation followed by the logistic function, but unlike before,
attention decisions are now conditioned on task-specific features.  We
use the following model (compare to Equation~\ref{eq:attention-study1}
from Modeling Study~1):
\begin{equation}\label{eq:attention}
a_i := \sigma\left(u + v^T {\hat w}_i + X_i^T A {\hat w}_i\right)
\end{equation}
As in Equation~\ref{eq:attention-study1}, $\hat w_i \in
\mathbb{R}^{100}$ is a word embedding representing $w_i$, and $u \in
\mathbb{R}$, $v \in \mathbb{R}^{100}$ are weight vectors.

The innovation compared to Equation~\ref{eq:attention-study1} is the
addition of the third term $X_i^T A {\hat w}_i$, which encodes
task-specific information.  $X_i \in \mathbb{R}^3$ is a feature vector
encoding the condition and, in the Preview condition, also whether the
token occurs in the question, detailed below.  The weight $A \in
\mathbb{R}^{3 \times 100}$ is a parameter of the model.

We build a single model that can simulate both experimental
conditions.  This way, the model can share parameters across
conditions, but we have to add some additional parameters indicating
differences between the conditions. Building separate versions of the
model for the two experimental condition would double the number of
parameters, making the model less parsimonious. It would also generate
two sets of results that are not directly comparable, making it hard
to evaluate whether the model shows the interactions we observed
experimentally in the eye-tracking data reported in
Section~\ref{sec:exp1} (especially the interactions with task,
which theoretically the most relevant).

The key to using a single model for both conditions is the feature
vector $X_i$, which encodes the following: (a)~the experimental
condition ($-0.5$ for Preview, $+0.5$ for No Preview), (b)~whether the
word occurs in the question, and (c)~an exponentially decaying running
average of~(b) over all fixated preceding tokens.

Crucially, the question feature~(b) (as well as feature~(c) derived
from it) is only available in the Preview condition, and zeroed out in
the No Preview condition.  This means that the attention module only
has information about the question in the Preview condition. In the No
Preview condition, the attention score $a_i$ has to be calculated
without taking the question into account.

Feature~(c) encodes the relevance of prior context to the question,
and represents a simple form of recurrence.
The decay factor for feature~(c) is a parameter of the model, learned
together with $u, v, A$.  \changedSecond{Feature (b) is} centered and scaled to
$[-0.5,+0.5]$ using the mean and range of the feature values estimated
from the training corpus.

\subsubsection{Task Module}
\label{sec:task_module}

Compared to Modeling Study~1, the most extensive changes are necessary
in the task module.  The task module of this version of NEAT performs
answer selection (recall that participants perform a multiple-choice
task in which they need to identify the correct answer among a set of
three distractors).

On a high level, similar to Modeling Study~1, the task module collects
the information gathered from the fixated words into a vector
representation, from which it then computes an answer.  For the
precise technical implementation of the module, we build on methods
from the recent natural language processing literature.  Specifically,
we build on the Attentive Reader model for textual question answering
described by \citet{hermann_teaching_2015}, with some simplifications
suggested by \citet{chen_thorough_2016}.  The input to the answer
selection module consists of the fixated words, and the neural
activations $h_i$ computed by the reader module for each word~$w_i$.
Crucially, skipped words are not provided to the task module, and
cannot be taken into account when answering the question.

\changedSecond{The task model first assembles all words read by the
  reader module using a bidirectional LSTM (BiLSTM), a standard neural network
  model for aggregating information from long sequences, consisting of
  two recurrent networks that aggregate the input in forward and
  backward direction.  For the forward component, we use the
  activations $h_i$ created by the reader; the backward component
  similarly creates a vector $r_i \in \mathbb{R}^{128}$ of neural
  activations for each token $w_i$.}

\changedSecond{Similarly, the question is summarized by a BiLSTM
networks, again with 128 memory cells each.}  Their final states are
concatenated to obtain a vector representation $r \in
\mathbb{R}^{256}$ for the question.  After that, for each token $w_i$
in the text, a number $b_i$ is computed as:
\begin{equation}\label{eq:reader-weights}
b_i = \exp\left(r^T B [h_i,w_i]\right)
\end{equation}
where $B \in \mathbb{R}^{256 \times 256}$ is a weight matrix and
$[h_i,r_i] \in \mathbb{R}^{256}$ is the concatenation of the vectors
$h_i$ and $r_i$.  The value of $b_i$ can be interpreted as encoding
the relevance of token $w_i$ to answering the question.

After that, the hidden states are averaged weighted with $b_i$ to
create a final vector representation as:
\begin{equation}
s := \frac{\sum_{i=1}^N b_i [h_i;r_i]}{\sum_{j=1}^N b_j}
\end{equation}
This representation, computed based both on the text and on the
question, is then used to choose the answer.  For this, a probability
distribution over the set of named entities is computed according to:
\begin{equation}\label{eq:reader-answer-selection}
t := \operatorname{softmax}(C \cdot s)
\end{equation}
where $C$ is a matrix, and $\operatorname{softmax}(x)_i :=
\frac{\exp(x_i)}{\sum_j \exp(x_j)}$ is an operation turning arbitrary
vectors into probability vectors.  The resulting $t$ is a vector of
probabilities, indexed \changed{by the set of named entities occurring in the text, numbered in the order of appearance. In order to indicate the indexing of named entities to the reader model, we allocate an embedding vector for every index up to the maximal number of named entities ($e_1, e_2, ..., e_{600}$), trained as model parameters, and add the relevant index embedding vector to the word embeddings when feeding it into the reader module.}
The probability vector in (\ref{eq:reader-answer-selection}) indicates how likely a given named entity is to be the correct answer to the question, given the text.\footnote{\changedSecond{It is important to note that the distractors (i.e., the incorrect answer choices) that are part of the task differ between the version of the task that participants perform, and the version of the task that the model performs. Unfortunately, this is unavoidable for practical reasons. For participants, we have to use distractors that are similar to the correct answer (e.g., if the correct answer is Michigan, then the distractors are also US states, see the example in Figure~\ref{fig:heatmap-human}). This is because using named entities from the text will make the task too easy. For the text in Figure~\ref{fig:heatmap-human}, this would results in a list of answers such as: Sabra, FDA, Michigan, March (these are all named entities in the text). For a human participant, it is obvious that Michigan is the correct answer. For the model, the situation is reversed: the model essentially has perfect memory, so it is trivial for it to determine that Michigan has appeared in the text, but other US states have not. On the other hand, for the model it is difficult to figure out which of the named entities in the text is a valid answer to the question it is asked. Therefore, non-answer named entities from the text are appropriate distractors for the model.}}

A question is counted as answered correctly if the named entity
corresponding to the correct answer is assigned a higher probability
than any other named entity in the probability vector~$t$. 
\changed{Our model has to choose among all named
entities occurring in the text, not just among the answer and the
three distractors. This means the task is considerably harder for our
model than for the participants in our eye-tracking study (a random
baseline would perform at $<10\%$ accuracy on most texts).
We note that it would technically possible to train the model to select from only four entities, but this would make the task almost trivial for the model, because it could easily check which one of the entities occurs in the text.}

\subsection{Objective Function}
\label{sec:neat_task:objective}

As in Modeling Study~1, we formalize the Tradeoff Hypothesis in terms
of an objective function weighting task success and the fraction of
fixated words.  However, task success is now defined not in terms of
reconstructing the input, but in terms of correctly answering
questions.

For a text $\boldsymbol{t}$ and a question $\boldsymbol{q}$ with
correct answer $a$, drawn from the corpus, NEAT stochastically chooses
a fixation vector $\boldsymbol{\omega} \sim
P_A(\boldsymbol{\omega}|T, \boldsymbol{t}, \boldsymbol{q}; \theta)$,
where $T \in \{\mathrm{Preview}, \mathrm{No Preview}\}$ is the
experimental condition.  Here, $\theta$ denotes a setting for the
parameters of the attention module.  The fixated words in the text,
together with the question, are passed to the answer selection module.
Success on question answering is formalized as the surprisal of the
correct answer $a$:
\begin{equation}
L(a|\boldsymbol{\omega}, \boldsymbol{t}, \boldsymbol{q}, T,\theta) := - \log P(a|\boldsymbol{\omega}, \boldsymbol{t}, \boldsymbol{q}, T,\theta)
\end{equation}
This term quantifies the loss on the question answering task and
should be \emph{minimized}.

As in the original NEAT model, we trade off task success and the
fraction of fixated words with a factor $\alpha > 0$. We average over
(a)~the two conditions, (b)~the texts and questions from the training
corpus, (c)~the fixation vectors $\boldsymbol{\omega}$ generated by
NEAT, to obtain the following loss function:
\begin{equation}\label{eq:objective:2}
	Q(\theta) :=  \frac{1}{2} \sum_{T \in \{\text{Preview, NoPreview}\}}  \operatorname{\mathbb{E}}_{(\boldsymbol{t}, \boldsymbol{q}, a)}
	\operatorname{\mathbb{E}}_{\boldsymbol{\omega}}\left[L(a|\boldsymbol{\omega}, \boldsymbol{t}, \boldsymbol{q}, T,\theta) + \alpha
    \cdot \frac{\|\boldsymbol{\omega}\|_{\ell_1}}{N} \right]
\end{equation}
Compared to Equation~\ref{eq:objective} in Modeling Study~1, the
differences are that (1)~loss is averaged across both conditions,
and (2)~task success is defined in terms of the probability assigned
to the correct answer, not reconstruction of the input.

We apply the same parameter estimation techniques as in Modeling
Study~1 (see Section~\ref{sec:parameters}), jointly training the
reader module, task module, and attention module with a combination of
gradient descent and reinforcement learning, to minimize the loss
function~(\ref{eq:objective:2}).

%%%%%%%%%%%%%%%%%%%%%%%%%%%%%%%%%%%%%%%%%%%%%%%%%%%%%%%%%%%%%%%%%%%%%%%%%%%%%%%%
\section{Modeling Study~2}
\label{sec:study2}

The aim of this modeling study is to test if the revised version of
NEAT that we developed in the previous section is able to correctly
account for the eye-tracking data of the task-based reading experiment
that we presented in Section~\ref{sec:exp1}. If NEAT is able to
capture the task variation attested in the experimental data,
then this will provide evidence for the Tradeoff Hypothesis that NEAT
is based on, and for our prediction that reading strategy is crucially
influenced by the task readers have to perform.

\subsection{Methods}
\label{sec:study2:methods}

\subsubsection{Model Implementation}

The reader module is implemented as an LSTM, in the same way as in the
original version of NEAT (see Section~\ref{sec:study1:methods}),
except that we now use a smaller hidden layer with only 128 memory
cells.  \citet{hermann_teaching_2015} and \citet{chen_thorough_2016}
showed that this lower dimensionality is sufficient for the question
answering task. Furthermore, the length of the texts prohibits the use
of larger layers due to memory limitations.

Following \citet{chen_thorough_2016}, the reader module uses pre-trained word embeddings from GloVe \citep{pennington_glove_2014}, which we continue training with the reader module.  The parameters in $\theta$ are initialized randomly
using the method described by \citet{glorot-understanding-2010}.

To choose $\alpha$ in~(\ref{eq:objective:2}), we ran the following
procedure. We first estimated parameters for each $\alpha$ from 0 to 2
in steps of size 0.2.  We then selected the $\alpha$ that resulted in
an overall fixation rate closest to the human fixation rate.  This way
we arrived at $\alpha = 1.0$.  The reinforcement learning algorithm
stochastically explores the action space and running it multiple times
may result in different strategies \citep{islam2017reproducibility};
this challenge is more pronounced in this second version of NEAT since
behavior depends on the task. In order to estimate the variance
introduced by this, we
ran the optimization algorithm 35 times to create 35 parameter
settings for $A$ for $\alpha = 1.0$.

The only ingredients of our model are a corpus of texts with questions
and answers, the neural architecture, and the objective function
in~(\ref{eq:objective:2}), combined with standard optimization
techniques for neural networks.  The reading strategies are optimized
on the basis of success in answering question.  As in Modeling
Study~1, no grammar, lexical knowledge, eye-tracking corpus, or other
labeled data are required to train our model.  \changed{Only one
  parameter, the scalar $\alpha$ in (\ref{eq:objective:2}), has to be
  adjusted to human eye-tracking data; all other parameters are chosen
  to optimize (\ref{eq:objective:2}).}

\subsubsection{Dataset}

We trained our model on the training section of the DeepMind question
answering corpus \citep{hermann_teaching_2015}, using the same corpus
pre-processing as the original authors.  This results in 1,259,748
text--question--answer triples.  \changed{On each pass through the
  training data, we downsampled the larger subcorpus (Daily Mail) so
  that an equal number of triples were sampled from the two sources
  (CNN and Daily Mail).}  We clipped long texts after 500 tokens due
to memory constraints.

We evaluated all model runs on \changed{the validation partition of
  the DeepMind corpus, where we downsampled the larger subcorpus as in
  training, resulting in 7,848 text--question pairs.}  For comparison,
we also ran the task module with all words fixated, which corresponds
to the original question-answering setup of
\citet{hermann_teaching_2015}.  \changed{We note that
  \citet{hermann_teaching_2015} represented named entities using
  anonymized identifiers (e.g., ``@entity7'' instead of
  ``London''). We reinserted named entities into the text as ordinary
  (capitalized) tokens; this makes the task harder for machine
  learning models, but matches the input available to the human
  participants.}

\subsection{Results}
\label{sec:study2:results}

\subsubsection{Tradeoff between Accuracy and Economy}

\begin{figure}[]
\begin{center}
  \includegraphics[width=.45\textwidth]{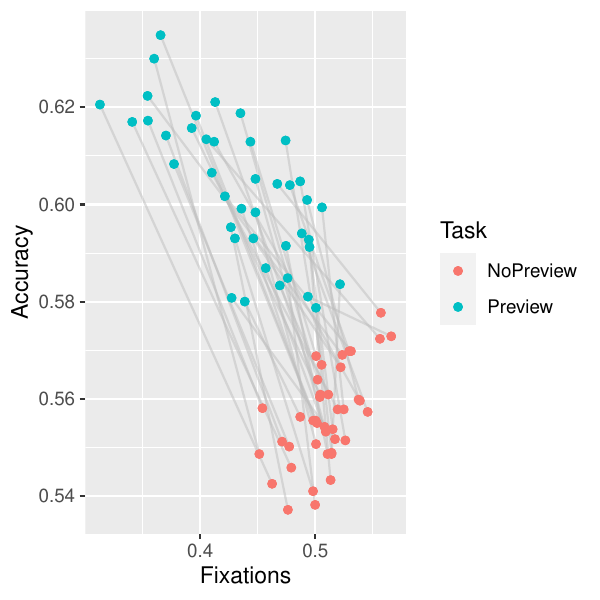}
\end{center}
	\caption{\changedSecond{Accuracy and economy in the two task
	conditions, at the selected tradeoff parameter ($\alpha=1.0$). Each pair of points linked by a gray line represents one run in the two task conditions. Across runs, the model tends to achieve higher accuracy and lower fixation rate in the Preview condition, when compared to the No Preview condition}. Numbers were computed on the validation partition of
  the DeepMind corpus (7,848 text--question
  pairs).
	\changedSecond{	See SI Appendix, Figure 3, for results across values of the tradeoff parameter $\alpha$.}
	}\label{fig:tradeoff}
\end{figure}

When all words are fixated, the task module of NEAT achieves a
question answering accuracy of 65.5\% -- similar to the accuracy of
66.1\% reported for the original model by
\citet{hermann_teaching_2015} on the same dataset, \changed{averaged
  across the CNN and Daily Mail sections}.  \changed{As noted above,
  \citet{hermann_teaching_2015} evaluated on an artificially modified
  task where named entities were anonymized, which we found makes the
  task easier for machine learning models.}

% Numbers referred to in this paragraph:
% ../study2/NEAT/results/output/fixationRates.txt
When we consider NEAT as a whole, i.e., including the reading module
and the attention module that skips words, we observe average fixation
rates of 51\% (SD 2.6\%) in the No Preview condition, and 44\% (SD
1.5\%) in the Preview condition.  NEAT therefore captures the
qualitative characteristics of the human data (see Table~\ref{tab:stat}): the
human fixation rate is 0.50 in the No Preview condition, and 0.34 (slightly
lower than predicted by the model) in the Preview
condition, close to what the model predicts.

Turning to question accuracy, we find that NEAT's accuracy is 56\% (SD
1.0\%) in the No Preview condition, and 60\%\% (SD 1.5\%) in the Preview
condition.  The model therefore shows the same qualitative effect as
in the human data: accuracy increases in the Preview condition, even
though fixation rate goes down.
Nevertheless, the model falls short of
human accuracy (which was 70\% in the Preview condition and 89\% in
the No Preview condition). However, the model has to choose between
all named entities appearing in the text, rather than just selecting one out
of four. That means the random baseline for human participants is
25\%, and for the model it is $<10\%$, i.e., the absolute accuracies
are not directly comparable (see Section~\ref{sec:task_module}).

In Figure~\ref{fig:tradeoff}, we plot accuracy as a function of
fixation rate. All models achieve higher accuracies with lower
fixation rate in the Preview condition.

% Result: ../study2/NEAT/train_NEAT/outputs/randomAttention.txt
For comparison, we also ran the question answering model with random
fixations at the same rate of 0.51 as in the No Preview condition. In
this setting, accuracy on question answering drops to 27\%, which
indicates that NEAT learns a skipping strategy which is clearly much
better than random even in the No Preview condition.

\begin{figure*}[]
\centering

No Preview

%{\scriptsize\input{heatmap-model-nopreview.tex}}
\vspace{2ex}
\includegraphics[width=.9\textwidth]{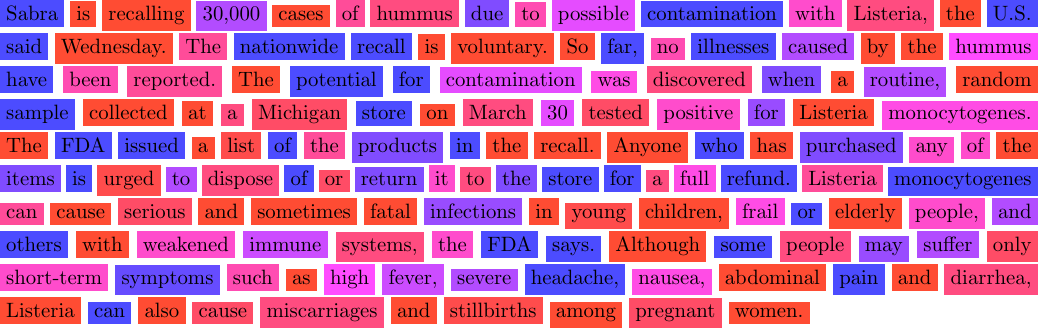}%

\vspace{2ex}
Preview

\vspace{2ex}

\includegraphics[width=.9\textwidth]{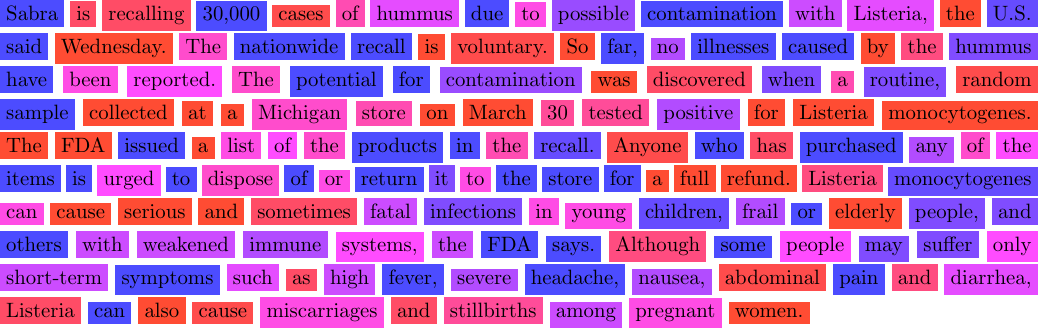}

\vspace{2ex}

\caption{Heatmap visualizing NEAT fixation probabilities in Modeling
  Study~2, averaged over 35 model runs. Top: No Preview, bottom:
  Preview.  The color gradient denotes the fixation probability
  predicted by NEAT for the word in a given condition, ranging from
  blue ($<0.3$) to red ($1.0$). The question and answer for this text
  are the same as in Figure~\ref{fig:heatmap-human}; the correct
  answer is the word \textit{Michigan} in the fourth line.  See SI
  Appendix, Figure 2 for a visualization of the difference between the
  two tasks.}
\label{fig:heatmap:model}
\end{figure*}

Figure~\ref{fig:heatmap:model} shows heatmaps visualizing NEAT
fixation probabilities for the same text as in
Figure~\ref{fig:heatmap-human} (which displays human fixation
probabilities).  This plot confirms that the overall fixation rate is
higher in the No Preview condition.  In both conditions, long words
and content words are more likely to be fixated.  In the Preview
condition, reading appears to be more targeted: Fixations are
concentrated on the sentence containing the answer \textit{Michigan}
in the fourth line.  The phrase \textit{tested positive for Listeria
  monocytogenes,} which appears in the question, shows higher fixation
rate in the Preview condition than in the No Preview condition.  In
other areas of the text, content words have decreased fixation
probabilities relative to the No Preview condition.

\subsubsection{Mixed Effects Analyses}

We ran each of the model parameterizations generated by the 35
training runs at $\alpha = 1.0$ on the same texts as used in
Experiment~1.

Then we built a logistic mixed effects model with NEAT fixation rate
as the dependent variable.  The random effects structure and the
priors were the same as when analyzing human fixation rate in
Experiment~1 (see SI Appendix, Section~2).  Our mixed effects model
includes model runs as a random effect; it plays a role analogous to
the random effect of participants in a mixed effects model of
experimental data.

As fixed factors, our mixed effects model included those that were
also used to analyze the eye-tracking data from Experiment~1 (see
Section~\ref{sec:exp1:methods}).  We built the final mixed model using
forward selection to include significant binary interactions into the
model, in the same way as for the eye-tracking
data.\footnote{\changedSecond{In Modeling Study~1, we used NEAT
    surprisal to predict human reading times. An analysis of this type
    doesn't make sense for Modeling Study~2, as the human reading data
    (see Table~\ref{tab:reading-times-lme:exp1}) did not show an
    interaction of surprisal and task, neither in the reading times
    nor in fixation rate. We therefore expect NEAT surprisal to be
    unaffected by task and instead analyze whether human fixation rate
    effects correspond to model fixation rate effects.}}

\begin{table}[]
\begin{center}
%\begin{tabular}{l@{}d@{}d@{~~}l@{~~}d@{}d@{~~}l@{~~}d@{}d@{~~}l@{~~}d@{}d@{~~~~}l@{}}
\begin{tabular}{l@{}r@{}r@{~~}r@{}r@{~~}r@{}r@{~~}r@{}r@{~~~~}}
\hline
& \mc{2}{c}{NEAT Fixation Rate}&  \\ \hline
(Intercept)                  &  \textbf{-0.19}  & (0.04) \\%& $^{}$ \\  
Task                    &  \textbf{0.25}  & (0.04) \\%& $^{***}$ \\ 
IsCorrectAnswer              &  -0.20  & (0.13) \\%& $^{}$ \\  
IsNamedEntity                &  \textbf{ 1.68}  & (0.10) \\%& $^{***}$ \\ 
WordLength                   &  \textbf{ 0.15}  & (0.01) \\%& $^{***}$ \\ 
LogWordFreq                  &  \textbf{-0.23}  & (0.01) \\%& $^{***}$ \\ 
PositionText                     &   0.02  & (0.02) \\%& $^{}$ \\  
Surprisal                    &   0.01  & (0.00) \\%& $^{}$ \\  
\hline
LogWordFreq:IsNamedEntity    &   \textbf{0.17}  & (0.01) \\%& $^{***}$ \\ 
LogWordFreq:WordLength       &   \textbf{0.02}  & (0.00) \\%& $^{***}$ \\ 
\hline
Task:IsNamedEntity      &   \textbf{-0.33}  & (0.05) \\%& $^{***}$ \\
Task:LogWordFreq        &   \textbf{-0.03}  & (0.01) \\% & $^{**}$ \\ 
%\multicolumn{13}{c}{$^{***}p<0.001$, $^{**}p<0.01$, $^*p<0.05$}
\hline
\end{tabular}
\vspace{1ex}
\caption{Logistic mixed effects model for fixation rate predicted by
  NEAT, with item and model run ($N=35$) as random effects.  Task was
  coded as $-0.5$ (Preview) vs. $+0.5$ (No Preview).
	\changedSecond{Binary interactions were identified automatically using forward model selection, as described in the text.}
	For each
  predictor, we give the coefficient and the standard
  deviation. Effects were interpreted as significant when the
  posterior probability that the coefficient has the opposite sign was
  < 0.05.}
\label{tab:mixed-neat}
\end{center}
\end{table}

\subsubsection{Task-independent Effects}

The result of the mixed effects analysis for NEAT fixation rate is
given in Table~\ref{tab:mixed-neat}. We will compare the significance
and the sign of the coefficients in our analysis of human fixation
rate in Table~\ref{tab:reading-times-lme:exp1}.

\changedSecond{We observe a significant, negative main effect of log
  word frequency and a significant, positive effect of word length on
  NEAT fixation rate. Both of these correspond to the effects found in
  human fixation rate. Furthermore, we observe a positive main effect
  of named entity status: words that are part of named entities are
  more likely to be fixated by NEAT. This effect is makes sense in
  terms of task adaptation: all the answers are named entities (in
  both the Preview and the No Preview condition), so fixating them
  more than normal words enables the model to perform more accurate
  question answering.  However, the effect of named entity status is
  not found in human fixation rate (though there is an
  IsNamedEntity:Surprisal interaction in human fixation rate which is
  not present in NEAT fixation rate).  NEAT shows no significant main
  effects of whether a word is part of the correct answer, its
  position in the text, or its surprisal; these effects are not
  present in human fixation rate either.}

\changedSecond{
  Turning now to the interactions, NEAT exhibits a significant
  interaction of log word frequency and whether a word is part of a
  named entity. We also find that log word frequency interacts with
  word length. Both interactions are present in the human fixation
  rate data. Human fixation rate also shows an interaction of named
  entity status and surprisal, which NEAT is not able to capture.}

\subsubsection{Task-dependent Effects}

We will now discuss main effect and interactions involving the factor
task, i.e., effects that depend on the reading task (Preview or No
Preview). Again, we compare the modeling results in
Table~\ref{tab:mixed-neat} with the human fixation data in
Table~\ref{tab:reading-times-lme:exp1}.

\changed{As we saw when discussing the descriptive statistics, NEAT
  fixation rate in the No Preview condition is higher than in the
  Preview condition. This is confirmed by a significant positive main
  effect of task in human fixation rate in
  Table~\ref{tab:mixed-neat}.}

\begin{figure}[]
\begin{center}
  \includegraphics[width=.45\textwidth]{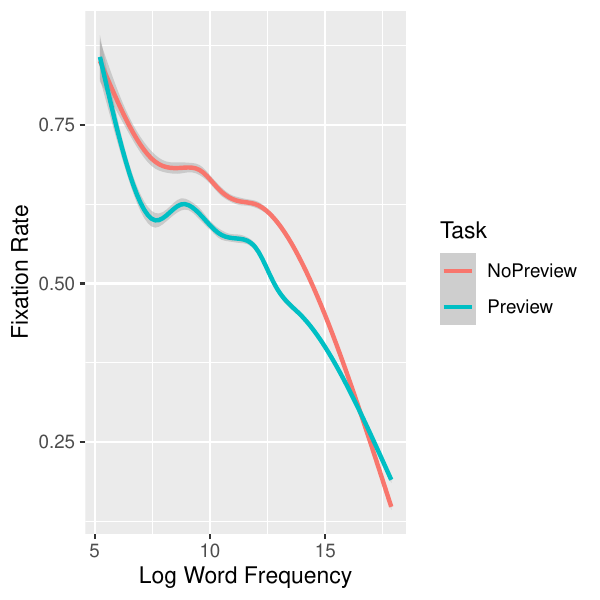}%
\end{center}
\caption{Model fixation rate by log word frequency by task condition.
  We show GAM fits with 95\% confidence intervals.  Compare
  Figure~\ref{fig:cond:logwf} for corresponding human data from
  Experiment 1.
	\changedSecond{See SI Appendix, Figure 4 for corresponding results including out-of-vocabulary tokens. } }\label{fig:model-wordfreq}
\end{figure}

\begin{figure}[]
\begin{center}
  \includegraphics[width=.45\textwidth]{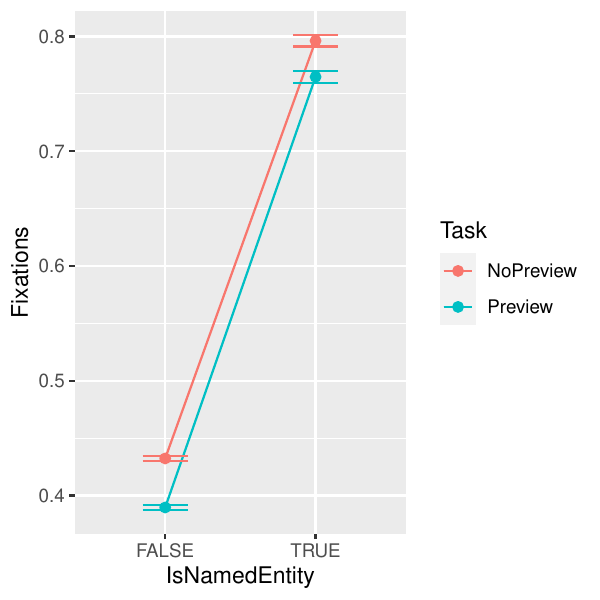}%
\end{center}
\caption{Model fixation rate by named entity status by task
  condition. Error bars indicate 95\% confidence intervals. Compare Figure~\ref{fig:cond:ne} for corresponding human
  data from Experiment 1.}\label{fig:model-isNE}
\end{figure}

\changedSecond{We also observe a significant negative interaction
  between task and log word frequency. This effect is illustrated in
  Figure~\ref{fig:model-wordfreq}.  The human data shows the same
  significant negative interaction in fixation rate. The explanation
  for this interaction is that NEAT is less guided by word frequency
  in its fixation decision when it knows which words to fixate, viz.,
  in the Preview condition, where it has read the question and knows
  what the answer should look like.  There furthermore is an
  interaction of task with named entity status, illustrated in
  Figure~\ref{fig:model-isNE}.  An effect of the same sign is also
  found in human fixation rate. This shows that NEAT is more likely
  to skip words that are part of named entities in the Preview
  condition than in the No Preview condition. This is an adaptive
  strategy, as in this condition, the model knows what type of named
  entity it is looking for (as it has seen the question), which allows
  it to spend less time on named entities overall.}

\changedSecond{Human fixation rate shows an interaction of task and whether
  a word is part of the correct answer or not. This interaction is not
  present in NEAT fixation rate; we will return to this in the
  discussion below. The interactions Task:Surprisal, Task:WordLength
  are not significant in both the human fixation rate and model
  fixation rate.}

\subsection{Discussion}
\label{sec:study2:discussion}

In this modeling study, we qualitatively evaluated the fixation
probabilities predicted by a new version of the NEAT reading model
which performs question answering. We conducted a mixed model analysis
on NEAT fixation rate and found that the effects mirrored the effects
found in human fixation rate in the eye-tracking experiment reported
in Section~\ref{sec:exp1}.  The results show that NEAT reading
behavior and human reading behavior are qualitatively similar:
Crucially, we found the same main effect of task, i.e., preview
affects fixation rate in the same way in the model and in the human
data. And we found that the experimental interactions between task and
word frequency and task and named entity status were also replicated
by the model.

One experimental effect that we failed to replicate was the
interaction of task and whether a word is part of the correct answer
or not. \changedSecond{This interaction was present in human fixation
  rate and there was an extraordinarily large effect in total time,
  but the interaction was not significant in NEAT fixation rate.  A
  possible explanation for the large effect in total time is that}
human readers exhibit a checking behavior in the preview condition
when they read a word that is part of the correct answer: in this
condition, they know what the answer should look like, so they re-read
it once they have encountered it in the text. This way they make sure
that they have really found the answer.  Our model has no way of
modeling regressions (as it's a model of skipping only).

%%%%%%%%%%%%%%%%%%%%%%%%%%%%%%%%%%%%%%%%%%%%%%%%%%%%%%%%%%%%%%%%%%%%%%%%%%%%%%%%
\section{General Discussion}
\label{sec:general_discussion}

In this article, we proposed NEAT, a neural network model of the
allocation of attention during human reading. NEAT is designed to
capture skipping, i.e., the process that decides which words in a text
should be fixated, and which ones should be skipped during
reading. NEAT is able to learn skipping strategies from large amounts
of text when given an explicit reading task, such as reconstructing
the input or answering questions about the text. The model is guided
by the Tradeoff Hypothesis: a successful reader trades off economy of
attention (skipping as many words as possible, i.e., reading as fast
as possible) and accuracy (making as few errors as possible in the
task the reader is trying to accomplish). The Tradeoff Hypothesis
predicts that task-specific reading strategies emerge when the
economy--accuracy tradeoff is optimized for a given task.

In Modeling Study~1, we implemented NEAT as a neural encoder-decoder
architecture with a hard attention mechanism and showed how the model
can the trained using reinforcement learning to optimize an objective
function that directly implements the Tradeoff Hypothesis. An
evaluation on the Dundee eye-tracking corpus
showed that NEAT predicts human fixation patterns through its measure
of fixation probability, and human reading time through its measure
of restricted surprisal.

A key prediction follows from the Tradeoff Hypothesis: reading
behavior is task-specific, as the tradeoff between economy and
accuracy can differ from task to task. Experiment~1 tested this
prediction in an eye-tracking experiment on newspaper text: In the No
Preview condition, participants read the text and then answered a
question about it. In the Preview condition, they first saw the
question, then read the text, and then answered the question. In the
Preview condition, participants read faster and skipped more, but
achieved higher answer accuracy compared to the No Preview
condition. Participants were also sensitive to the task relevance of
words: In the No Preview condition, they spent more time reading words
that were part of named entities (which are potential answers); in the
Preview condition, they instead spent more time on words that were
part of the answer (checking that they have found the answer). This
provides evidence that reading strategy depends on whether
participants perform a task that is similar to standard reading (No
Preview), or an information seeking task (Preview).

The aim of Modeling Study~2 was to design and evaluate a computational
model that captures the reading behavior observed in Experiment~1. We
achieved this by developing a modified version of NEAT which incorporated a
task module that performs question answering. We analyzed the skipping
behavior of the revised model and found that it predicted both the
main effect of task condition and the interactions of task with word
frequency and named entity status observed in Experiment~1.  This
indicates that NEAT is able to develop task-based reading strategies
that mimic those found in human readers.

Taken together, our modeling studies showed that NEAT successfully
captures human skipping behavior during the reading of text. In
particular, the model is able to change its reading strategy to
accomplish a particular task, in line with what humans do in
task-based reading. Crucially, the behavior of our model emerged when
we combined a neural network architecture that is designed to
accomplish a key aspect of reading behavior, viz., deciding whether a
word should be fixated or not, with a task-based objective, such as
reconstructing the content of a text, or answering a question about
the text. There was no need to explicitly include task-relevant
features (such as word frequency or named entity status) into the
model. Our model learns to pay attention to such features when trained
on a large collection of texts (and question-answer pairs in Modeling
Study~2), using only general optimization strategies such as
reinforcement learning and backpropagation of errors.

\subsection{Limitations}

In future work, we aim to address some key limitations of our model.
An important aspect of reading is that it in fact operates not on
the word level, but on the character level. Human readers target their
fixations at a specific character within the word, and then process
information within a 7--9 character window. Depending on where they
land on a word, a human reader may gain only a partial view of the
word they are fixating, or they may be able to view some characters of
an upcoming word. NEAT, which treats words as fixed-length word
embeddings, is not able to capture landing position or preview
effects. This is something that could be remedied by switching from
word embeddings to character embeddings, an approach that has been
very successful in the natural language processing literature
\cite[e.g.,][]{kim16}, and by modeling surprisal on the level of
character input \citep{Hahn:ea:19}.

More generally, our model only describes skipping of words (and
computes skipping-based surprisal).  Real eye-movements during reading
are substantially more complex: a word can be fixated more than once
in first-pass reading, or the word can be refixated once it has been
left, either in a forward saccade or in a regression. Punctuation
marks, phrase and sentence boundaries influence eye-movements, and so
do end of lines, which necessitate specific return sweeps. Sometimes
reading difficulty spills over to the next word, and causes increased
reading time there.  All of these phenomena should be captured by a
comprehensive model of human reading.  The coverage of our model could
be extended by incorporating features of models that provide accounts
at the level of saccade programming, such as
E-Z Reader~\citep{reichle_toward_1998, reichle_ez_2003,
  reichle_using_2009} and SWIFT~\citep{engbert_dynamical_2002,
  engbert_swift:_2005}.

\subsection{Relation to other Models of Reading}
\label{subsec:other-models}

Our model focuses on the allocation of attention and how it is
influenced by the task, whereas models such as E-Z Reader
\citep{reichle_toward_1998, reichle_ez_2003, reichle_using_2009},
SWIFT \citep{engbert_dynamical_2002, engbert_swift:_2005}, and
OB1~Reader \citep{snell2018ob1} provide accounts of the programming
and execution of saccades at the character level.  Our model does not
aim to capture reading at the same level of detail as those models. It
does not explicitly model saccades at the character level, nor provide
a detailed account of word length and word frequency effects. It is
also unable to capture regressions (reverse eye-movements).

Another difference between our model and E-Z Reader and SWIFT is
that NEAT comes with an unsupervised learning method. We only have to
specify how the reader trades off economy and accuracy, represented by
the factor $\alpha$ in equation~(\ref{eq:objective}). We do not need
to fit any model parameters directly on eye-tracking data. Through the
use of reinforcement learning, NEAT relies on a cognitively plausible
learning method that can discover efficient reading strategies based
on how successful they are for a given task.

% Proposed new text -- Looks good, FK
\changedSecond{It is conceivable that aspects of NEAT can be
  integrated into existing models of eye-movement control. For example
  E-Z Reader assumes a component that carries out a familiarity check
  on the word currently being processed. The duration of this check is
  modulated by the word's frequency and its predicability within the
  sentence. In E-Z Reader, frequency and predicatbility (cloze
  probability) are pre-specified constants. NEAT on the other hand,
  assumes that predicability that is learned from experience (e.g.,~by
  training the model on a text corpus). It may therefore be possible
  to augment E-Z Reader with a NEAT-style model of predictability
  (which would also naturally incorporate corpus frequency).}

\changedSecond{Alternatively, NEAT could be extended to model reading
  at the level of eye-movement control.  For instance, if we enhance
  NEAT with a character-level language model, then it is able to
  predict on which character a fixation will land. This approach has
  been successfully implemented by \citet{Yan:ea:22}, who show that
  even though it is trained only on unannotated text, their version of
  NEAT provides fit to human fixation positions competitive with E-Z
  Reader.} Besides modeling input at the level of characters,
\citet{bicknell_rational_2010-1} also suggest a way of capturing
realistic visual input including uncertainty. This could be integrated
into our model by conditioning the attention module $A$ on noisy
visual input.  \changedSecond{More ambitiously, future work might
  develop neurally-parameterized models of eye-movement control at the
  level of detail of SWIFT or E-Z Reader, and embed them within NEAT
  in place of~$A$, in order to account for how reading behavior is
  influenced by everything from low-level oculumotor control to
  high-level task effects (as investigated in this paper).}

Our proposed Tradeoff Hypothesis follows a tradition of modeling
reading as rational behavior.  Prior rational models have focused on
word identification, finding optimal solutions to identifying the
words in a sentence in the minimum number of saccades and/or the
minimum amount of time \citep{legge1997mr, legge2002mr, nor06,
  reichle2006using, bicknell_rational_2010-1, Lewis:ea:13}.  Several
models \citep{legge2002mr, nor06, bicknell_rational_2010-1,
  Lewis:ea:13} define policies that aim to minimize uncertainty about
the current word before moving to the next word (or make a
regression).  The models of \citet{reichle2006using} and
\citet{Lewis:ea:13} are particularly related to ours in that they
optimize policies for rewards that explicitly trade off economy with
accuracy (on word identification, \citealp{reichle2006using}, or
lexical decision, \citealp{Lewis:ea:13}).  Beyond language, models of
visual behavior based on reinforcement learning have been proposed in
other domains \citep[e.g.,][]{Sprague2007ModelingEV,
  NuezVarela2013ModelsOG, Hayhoe2014ModelingTC, Butko2008IPOMDPAI,
  Acharya2017HumanVS}, using both policy-gradient methods like in our
model \citep{Butko2008IPOMDPAI} and Q-Learning algorithms
\citep{Sprague2007ModelingEV, NuezVarela2013ModelsOG,
  Acharya2017HumanVS}.

Compared to earlier rational models of reading, NEAT innovates in two
main respects: On a technical level, while these earlier models
assumed restricted state spaces with manually constructed features,
and a closely delimited input domain of word lists, the combination of
reinforcement learning with neural network modeling enables NEAT to be
applied to real-world text data, without providing the model with
information about what features (e.g.,~word frequency) to pay
attention to.  On a theoretical level, NEAT expands the domain of
rational modeling of reading to language understanding tasks beyond
word identification, providing a theoretically motivated account of
task effects in high-level tasks such as question answering.

\changed{Of particular relevance to ours is the model of
  \citet{bicknell_rational_2010-1}. Compared to this model, NEAT
  offers advances in several respects. On the theoretical level, NEAT
  gives an account of task-dependent reading: Whereas
  \citet{bicknell_rational_2010-1} focused on word identification,
  NEAT provides a general theory of task-specific reading behavior,
  including a plausible learning algorithm based on reinforcement
  learning. On a technical level, thanks to the use of contemporary
  neural network modeling, NEAT is scalable to arbitrary input and
  high-level language understanding tasks with little task-specific
  adaptation to the model. Similar arguments apply to the related
  model proposed by \citet{Lewis:ea:13}, which is assumes a fixed task
  and a predefined set of payoff schemes to model tradeoff between
  speed and accuracy. Also, this model is only developed and tested
  for a list lexical decision task. It is not clear how this approach
  can capture effects across tasks, or where no explicit payoff is
  given.}

As described in Section~\ref{sec:eye-movement_models}, there is also a
line of work treating the prediction of eye-tracking measures as a
supervised learning problem.  One aim of this literature is to improve
natural language processing systems by making their attention patterns
similar to human attention patterns using supervised learning
\citep{DBLP:conf/conll/BarrettBHRS18, Barrett2020SequenceLA}.  Most
closely related to our work, \citet{malmaud2020bridging} and
\citet{DBLP:conf/conll/SoodTFBV20} collected eye-tracking data in
reading comprehension, and showed improvements in the accuracy of
automated question answering by training models to predict human
reading measures \citep{malmaud2020bridging,
  DBLP:conf/nips/SoodTMB20}.  \changedSecond{A recent machine
  learning-based approach that treats eye-movement prediction as an
  unsupervised learning task is presented by
  \citet{Yang2022UnsupervisedTS}. These authors are able to show that
  text segmentation algorithms are predictive of fixation locations.}

\changed{Besides fixation probabilities, NEAT computes a version of
  surprisal based on the fixated words only. It has some resemblance
  to Lossy Context Surprisal
  \citep{Futrell2020LossyContextSA}. However, the two notions are
  conceptually distinct, because Lossy-Context Surprisal assumes that
  the relevant limitations arise in memory representations and memory
  retrieval, whereas NEAT surprisal models limitations that arise from
  the reading process.}

\subsection{Theoretical Import and Relation to Speed-Accuracy
  Tradeoff}

\changed{Speed-accuracy tradeoffs are common in human decision making
  (see \citealt{Heitz:14}, for an overview), so assuming that they
  also explain the task adaptiveness of human reading may seem
  self-evident. \changedSecond{However, none of the work reviewed in
    Section~\ref{sec:tradeoff} conceptualizes task effects in reading
    in terms of a speed-accuracy tradeoff.} Instead, a large number of
  unrelated theoretical concepts are assumed:
  \citet{Rayner:Fischer:96} and \citet{Radach:ea:08} compare normal
  reading, scanning, and word verification and infer the existence of
  different ``modes'' of eye-movement control, which they attribute to
  top-down influences on semi-autonomous cognitive modules.
  \citet{Greenberg:ea:06} study letter detection tasks during reading
  and attribute task effects to differences in word class
  processing. \citet{White:ea:15} also study scanning and normal
  reading, and attribute task differences in post-lexical integration
  (i.e.,~the formation of a coherent text representation). Studying
  proofreading vs. reading for comprehension,
  \citet{Kaakinen:Hyona:10} explain task differences in terms of
  variations in perceptual span. \citet{Schotter:ea:14} also study
  proofreading and present a more elaborate theoretical framework,
  which assumes five component processes central in normal reading:
  wordhood assessment, form validation, content access, integration,
  and word-context validation. They hypothesize that different types
  of proofreading emphasize specific component processes, thus
  explaining task differences.  \citet{Kaakinen:ea:15} compare reading
  for question answering with reading for comprehension and attribute
  task difference to divergent memory representations being
  constructed for different tasks.}

\changed{The ambition of our work is to offer a unified idea that can
  bring together this diverse set of theoretical assumptions. We
  conjecture that the task effects found in the experimental
  literature can be explained by a tradeoff between economy of
  attention and task accuracy, provided we have a model of attention
  selection and a task model, and we combine these with a suitable
  optimization scheme. In this article, we showed that the NEAT model
  in conjunction with reinforcement learning can achieve this for a
  small number of example tasks (text reconstruction; question
  answering with and without preview). Our results therefore provide a
  case study that can serve as a first step towards the unification of
  existing diverse explanations for task effects in reading.}

\changed{It is also important to point out that the speed-accuracy
  tradeoff literature, while conceptually similar to our approach,
  differs in an important aspect: It typically assumes a single task
  and manipulates task payoff to obtain a tradeoff between speed and
  accuracy for that task \citep[e.g.,][]{Lewis:ea:13}. Our work, in
  contrast, shows that we can predict difference in reading behavior
  across tasks if we assume a realistic computational model of the
  tasks themselves. We can then derive the payoff (the reward in our
  reinforcement learning setup) from the model, rather than having to
  pre-specify it externally. This highlights the theoretical import of
  the work: given the right learning mechanism and enough data
  (e.g.,~texts, question-answer pairs), we can learn the tradeoff
  between economy and accuracy, without having to specify external
  payoffs. (Such external payoff are not a realistic assumption in
  naturalistic reading, they are typically found in a lab setting
  only.)}

\changed{In more general terms, our model offers not only a simulation
  of reading behavior, it does so by assuming a cognitively plausible
  architecture together with learning mechanism that uses realistic
  input (real text), while also respecting developmental constraints
  (e.g.,~we do not assume that eye-movement data are available during
  learning).  This contrasts with previous models of eye-movement
  control such as E-Z Reader and SWIFT which presuppose that certain
  factors (e.g.,~word frequency, word length, predictability) affect
  reading behavior.  Our model shows that these factors do not need to
  be stipulated: Not only can their influence be learned from raw
  text, but the model also learns how they interact with task
  factors.}

\subsection{Relation to Task Effects in the Literature}

Our results in Experiment~1 add to the literature on task effects in
human reading behavior (see Section~\ref{sec:tradeoff} for a review).
We found strong main effects of the task manipulation in both reading
time and fixation rate, with faster reading when a question preview
was available.  Main effects of task manipulations have been observed
in many of the prior studies on task effects.  Most closely related to
our results, it has been found that manipulating the type or
difficulty of questions changes reading measures, such that harder
questions lead to higher reading time and more
fixations~\citep{Radach:ea:08, wotschack2013reading,
  weiss2018comprehension}.

We also found interactions of the task manipulations and other
predictors.  Effects of word frequency and word length were reduced in
the NoPreview condition.  Task-dependent modulation of frequency
effects has also been reported in previous
studies~\citep{Radach:ea:08, Kaakinen:Hyona:10, wotschack2013reading,
  Schotter:ea:14}.  While some task manipulations have been shown to
also interact with predictability~\citep{wotschack2013reading,
  Schotter:ea:14}, surprisal and condition did not interact in our
Experiment~1.

We further found interactions with predictors specific to the task,
namely IsCorrectAnswer and IsNamedEntity.  In agreement with this
finding, \cite{malmaud2020bridging} showed -- in an experiment
building on our Experiment~1 -- increased reading time in regions
critical for answering a question when a preview of the question was
available, analogous to the interaction between condition and
IsCorrectAnswer we found in Experiment~1.

%%%%%%%%%%%%%%%%%%%%%%%%%%%%%%%%%%%%%%%%%%%%%%%%%%%%%%%%%%%%%%%%%%%%%%%%%%%%%%%%
\section{Conclusions}
\label{sec:conclusions}

We introduced NEAT, a model of the allocation of attention in human
reading.  NEAT is implemented using neural network-based attention, and is trained
using reinforcement learning to trade off economy of attention with
task accuracy.  We showed that NEAT predicts human fixation patterns
and readings times on the Dundee Corpus.  A key prediction of NEAT is
that the allocation of attention should adapt to the task.  We tested
this in an eye-tracking experiment, finding that reading behavior in a
reading comprehension task differed depending on the availability of a
preview of the questions that readers had to answer.  We showed that
the experimentally attested task differences were predicted
by a version of NEAT trained for
question answering, lending support to the Tradeoff Hypothesis.  Taken
together, our results support NEAT as a model of attention in human
reading, and suggest more generally that task effects in human reading
reflect efficient adaptations to the properties of tasks.

\section*{Acknowledgments}

We gratefully acknowledge the support of the Leverhulme Trust (award
IAF-2017-019 to FK).  We are also grateful to Chris Manning, Dan
Lassiter, members of the Stanford NLP Group, and the CUNY 2017
audience for helpful feedback.

%%%%%%%%%%%%%%%%%%%%%%%%%%%%%%%%%%%%%%%%%%%%%%%%%%%%%%%%%%%%%%%%%%%%%%%%%%%%%%%%
\renewcommand{\bibsection}{\section*{References}}

\bibliographystyle{apa}
\bibliography{references}

\end{document}